\documentclass[journal]{IEEEtran}
\IEEEoverridecommandlockouts

\usepackage{amsmath,mleftright,mathtools,amsthm}
\usepackage{amssymb}
\usepackage{lipsum}
\usepackage[pdftex, pdfstartview={FitV}, pdfpagelayout={TwoColumnLeft},bookmarksopen=true,plainpages = false, colorlinks=true, linkcolor=black, citecolor = black, urlcolor = black,filecolor=black , pagebackref=false,hypertexnames=false, plainpages=false, pdfpagelabels ]{hyperref}
\usepackage{balance}
\usepackage{xargs}
\usepackage{graphicx}
\usepackage[labelformat=simple]{subcaption}
\DeclareCaptionLabelSeparator{periodspace}{.\quad}
\usepackage{enumitem}
\usepackage[sort,compress]{cite}

\usepackage{afterpage}
\usepackage{tabularx}
\usepackage{bm,etoolbox}
\usepackage[linesnumbered,ruled,vlined]{algorithm2e}
\usepackage[pdftex,dvipsnames]{xcolor}
\usepackage{rotating}
\usepackage[makeroom]{cancel}

\usepackage{dblfloatfix}
\DeclareGraphicsExtensions{.pdf,.png}
\graphicspath{ {./figs/} }
\DeclarePairedDelimiterX{\infdivx}[2]{(}{)}{%
  #1\;\delimsize\|\;#2%
}

\newcommandx{\unsure}[2][1=]{\todo[linecolor=red,backgroundcolor=red!25,bordercolor=red,#1]{#2}}
\newcommandx{\add}[2][1=]{\todo[linecolor=blue,backgroundcolor=blue!25,bordercolor=blue,#1]{#2}}
\newcommandx{\info}[2][1=]{\todo[linecolor=OliveGreen,backgroundcolor=OliveGreen!25,bordercolor=OliveGreen,#1]{#2}}
\newcommandx{\edit}[2][1=]{\todo[linecolor=Plum,backgroundcolor=Plum!25,bordercolor=Plum,#1]{#2}}
\newcommandx{\hide}[2][1=]{\todo[disable,#1]{#2}}

\SetCommentSty{mycommfont}

\usepackage{accents}  

\DeclareMathOperator{\SO}{SO}
\newcommand{\so}{\mathfrak{so}}

\DeclareMathOperator*{\argmin}{argmin}

\theoremstyle{definition}

\def \latVar{\Theta}
\def \latVars{\boldsymbol{\Theta}}

\def \latVals{\boldsymbol{\theta}}
\def \measVal{z}
\def \measVals{\mathbf{z}}
\def \measVar{Z}

\def \clique{C}
\def \separator{S_\clique}
\def \frontal{F_\clique}

\newcommand{\children}[1]{\mathcal{C}_{#1}}
\newcommand{\parents}[1]{\Pi_{#1}}
\def \pClique{\parents{\clique}}
\def \cClique{\children{\clique}}

\def \allFactors{\mathbf{f}}

\def \allE{\mathcal{E}}

\newcommand{\leftSlice}[2]{{#1}_{<#2}}
\newcommand{\rightSlice}[2]{{#1}_{>#2}}

\newcommand{\bigCI}{\mathrel{\text{\scalebox{1.07}{$\perp\mkern-10mu\perp$}}}}

\newcommand{\condIndep}[3]{{#1}\bigCI{#2}|{#3}}

\newcommand{\sampling}[1]{{#1}\sim{p({#1})}}

\def \bigdot{\boldsymbol{\cdot}}
\theoremstyle{plain}

\theoremstyle{definition}

\theoremstyle{remark}

\title{\LARGE \bf
Incremental Non-Gaussian Inference for SLAM\\ Using Normalizing Flows
}

\author{Qiangqiang Huang$^{1,*}$, Can Pu$^{2}$, Kasra Khosoussi$^{3}$, David M. Rosen$^{4}$, Dehann Fourie$^{5}$,\\
Jonathan P. How$^{6}$, and John J. Leonard$^{1}$
\thanks{$^{1}$MIT Computer Science and Artificial Intelligence Laboratory, Cambridge, MA 02139, USA. {\tt\{hqq,jleonard\}@mit.edu} *Corresponding author.}%
\thanks{$^{2}$MIT Department of Nuclear Science and Engineering, Cambridge, MA 02139, USA. {\tt\small {pucan}@mit.edu}}%
\thanks{$^{3}$Robotics and Autonomous Systems
Group, DATA61, CSIRO, Brisbane, QLD 4069, Australia. {\tt\small kasra.khosoussi@data61.csiro.au}}%
\thanks{$^{4}$Northeastern University Department of Electrical and Computer Engineering, Boston, MA 02115, USA. {\tt\small{d.rosen}@northeastern.edu}}%
\thanks{$^{5}$NavAbility, Boston, MA 02110, USA. {\tt\small{dehann}@navability.io}}%
\thanks{$^{6}$MIT Department of Aeronautical and Astronautical Engineering, Cambridge, MA 02139, USA. {\tt\small jhow@mit.edu}}%
\thanks{Research supported by ONR grant N00014-18-1-2832 and ONR MURI grant N00014-19-1-2571.}
}

\begin{document}


\maketitle
\thispagestyle{plain}
\pagestyle{plain}

\begin{abstract}
This paper presents normalizing flows for incremental smoothing and mapping (NF-iSAM), a novel algorithm for inferring the \emph{full} posterior distribution in SLAM problems with nonlinear measurement models and non-Gaussian factors. NF-iSAM exploits the expressive power of
neural networks, and trains normalizing flows to model and sample the full posterior. By
leveraging the Bayes tree, NF-iSAM enables efficient incremental
updates similar to iSAM2, albeit in the more challenging \emph{non-Gaussian} setting. We demonstrate the advantages of NF-iSAM over state-of-the-art point and distribution estimation algorithms using range-only SLAM problems with data association ambiguity. NF-iSAM presents superior accuracy in describing the posterior beliefs of continuous variables (e.g., position) and discrete variables (e.g., data association).
\end{abstract}


\section{Introduction}
\label{sec:introduction}

Simultaneous localization and mapping (SLAM) is a foundational capability for mobile robots, enabling such basic functions as planning, navigation, and control.  As such, the development of \emph{robust}, \emph{accurate}, and \emph{computationally efficient} SLAM algorithms has been a major focus of research in robotics over the previous three decades \cite{durrant2006simultaneous, Cadena2016SLAM,Rosen2021Advances}. 

We state the estimation problem for SLAM as inferring the \emph{full posterior distribution} of latent variables (i.e., robot and landmark poses) provided noisy relative measurements between those variables. Note that the distribution estimation is different from the point estimation typically seen in the SLAM literature (e.g., the MAP estimation in \cite[Fig. 2]{Cadena2016SLAM}). We pursue the full posterior distribution since estimates of the distribution are required in many applications including probabilistic data association, collision avoidance, and active perception.

\begin{figure}[t!] 
 	\centering
 	\includegraphics[width=0.95\columnwidth]{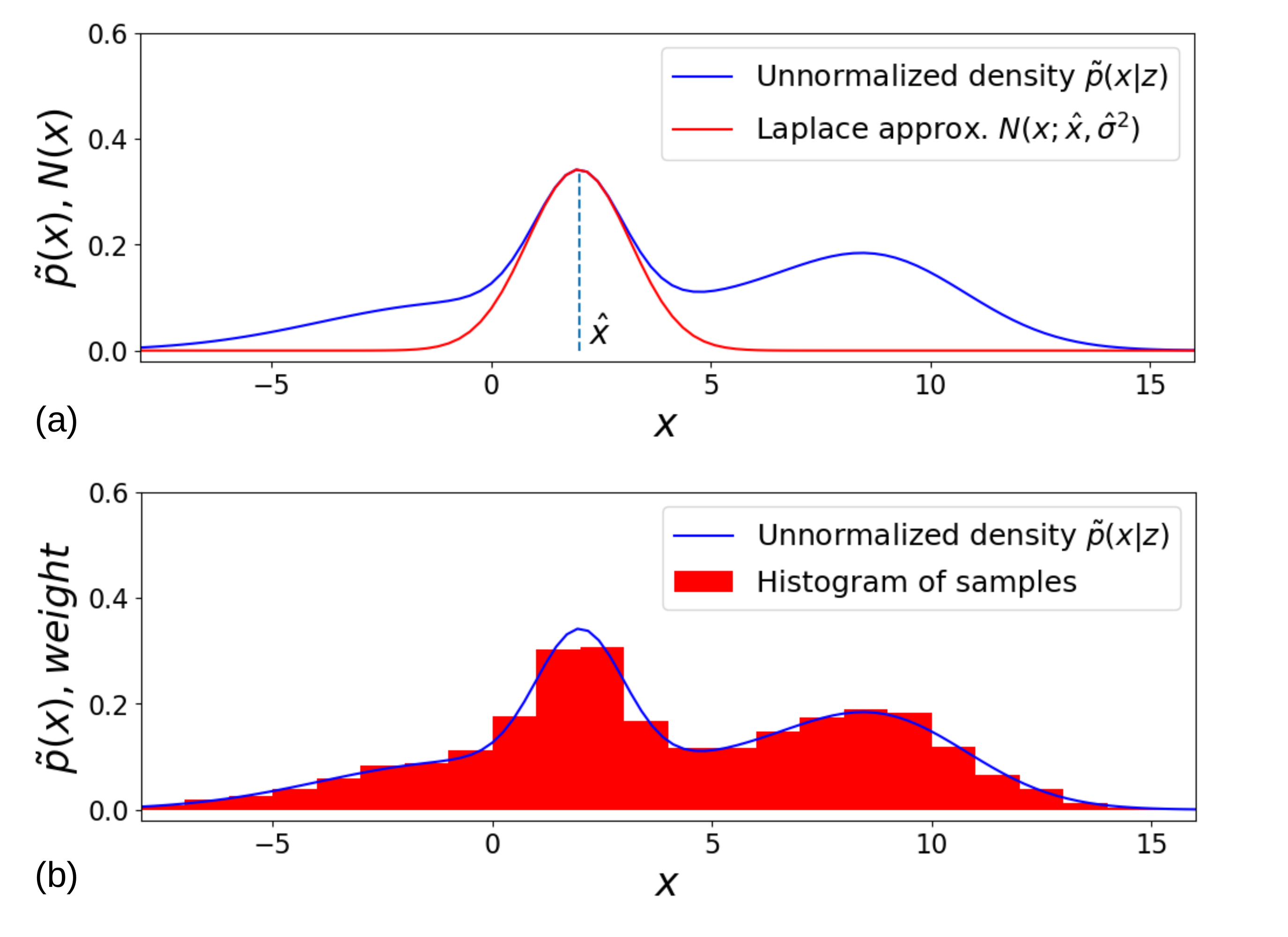}
     \caption{Approximate distributions of an unnormalized posterior distribution, $\tilde{p}(x|z)$: (a) the Laplace approximation centering on the MAP estimate, $\hat{x}$, and (b) Monte Carlo approximation.}
     \vspace{-0.5 cm}
     \label{fig: two methods}
\end{figure}

Current state-of-the-art SLAM algorithms (such as iSAM2~\cite{kaess2012isam2}) seek to recover a maximum \emph{a posteriori} (MAP) estimate using nonlinear local optimization~\cite{boyd2004convex}. This approach is attractive because using sparsity-exploiting first- or second-order optimization methods permits fast recovery of the MAP (i.e. \emph{point}) estimate. Furthermore, assuming that the true posterior distribution is highly concentrated around the point estimate, one can construct a Gaussian approximation to the \emph{full} posterior by applying the Laplace approximation (see Fig.~\ref{fig: two methods}a) \cite[Ch. 4.4]{bishop2006pattern}. However, the posterior distribution in real-world SLAM problems is often non-Gaussian and may have multiple modes. This is in part due to non-linear measurement models and non-Gaussian factors \cite{rosen2013robust,olson2013inference}. Real-world examples include systems with range measurements \cite{blanco2008pure}, pose transformations on the special Euclidean group \cite{long2013banana}, multi-modal data association~\cite{doherty2019multimodal}, the (bimodal) slip/grip behavior of odometry measurements~\cite{thrun2002probabilistic}, multipath effects of sonar and radar~\cite{sunderhauf2013switchable}, and object pose ambiguity in images due to occlusion and symmetry \cite{lu2021consensus,pmlr-v139-murphy21a,deng2021poserbpf, fu2021multi}. Therefore, the use of a Gaussian (or any other unimodal) model of posterior is \emph{inherently} incapable of capturing critical information about the true uncertainty in SLAM posteriors, which is essential for safe navigation.

As an alternative to (unimodal) optimization-based MAP (i.e.\ \emph{point}) estimators, one might consider approximating the full posterior using nonparametric (e.g.\ sample-based, see Fig.~\ref{fig: two methods}b) methods for greater expressive power. The main challenge with this approach is that estimating the full SLAM posterior is at least NP-hard in general \cite{Rosen2021Advances}. As a result, the computational cost of standard general-purpose Bayesian inference methods, such as Markov chain Monte Carlo (MCMC) or nested sampling, is usually too large for the high-dimensional posteriors that arise in SLAM applications \cite{torma2010markov,skilling2006nested, huang2021reference}. Some recent efforts have attempted to improve the computational speed of nonparametric methods by exploiting the conditional independence structure in SLAM factor graphs~\cite{fourie2017multi,hsiao2019mh}. One such approach is to extend the Bayes tree, which was proposed in iSAM2 for analyzing the Gaussian approximation, to non-Gaussian settings~\cite{kaess2012isam2}. The Bayes tree algorithm converts a cyclic factor graph into an acyclic directed graph using a variable elimination game~\cite{Heggernes1996finding} and max-cardinality search~\cite{Tarjan1984simple}. From an information-theoretic standpoint, the Bayes tree shows how to factorize the original \emph{high-dimensional} posterior into a \emph{sequence} of \emph{low-dimensional} conditionals that encode a tree-like graphical model. Recent multi-modal extensions of iSAM2, such as mm-iSAM~\cite{fourie2017multi} and MH-iSAM2~\cite{hsiao2019mh}, all take advantage of the acyclic Bayes tree
to solve the original inference problem by performing inference over the decomposed lower-dimensional problems. These algorithms, however, can only infer the \emph{marginal} posterior distribution (mm-iSAM) or are limited to certain sources of non-Gaussianity (MH-iSAM2); in addition, none of these algorithms explicitly model non-Gaussian conditionals in the Bayes tree.


The technical goal in this paper is to find a computationally tractable density representation that has the necessary flexibility to approximate the full posterior distribution.  Specifically, we aim to develop an algorithm that is able to perform the following tasks in non-Gaussian settings:
\begin{enumerate}
    [leftmargin=*,
    label={\textit{Task}\ \arabic*.},
    ref={\textit{Task}\ \arabic*}]
    \item   Solve for a distribution that effectively approximates the full posterior\label{task-1}.
    \item   Draw samples from the distribution to infer quantities of interest using Monte Carlo integration\label{task-2}.
    \item   Allow incremental updates of the distribution\label{task-3}.
\end{enumerate}
We propose to learn \emph{non-Gaussian} models of conditionals that factorize the posterior using the Bayes tree. The learned non-Gaussian models in turn reconstruct the posterior. To perform inference, we draw samples sequentially from these models following the order governed by the Bayes tree.
We exploit normalizing flows to represent the non-Gaussian conditionals. Normalizing flows, as emerging tools for density modeling~\cite{rezende2015variational, durkan2019neural,jaini2019sum,rezende2020normalizing,kobyzev2020normalizing,papamakarios2021normalizing},  have shown strong expressive power for representing complex densities and support fast sampling. An important property is that conditionals of the modeled density can be extracted easily from a normalizing flow model, which perfectly matches our need for modeling conditionals.

\subsection*{Contributions}
We present a novel general solution, called normalizing flows for incremental smoothing and mapping (NF-iSAM)\footnote{NF-iSAM is open source at \url{https://github.com/MarineRoboticsGroup/NF-iSAM.git} including source code and the datasets used in this paper.}, to model and sample the \emph{full posterior distribution} of general SLAM problems using the Bayes tree and the normalizing flow model. Key contributions of this work include:
\begin{enumerate}
    \item  NF-iSAM introduces normalizing flows to factor graph inference for robot perception.
    \item  NF-iSAM generalizes the Bayes tree to perform full (non-Gaussian) posterior estimation for the joint posterior distribution.
    \item  NF-iSAM augments normalizing flows from low-dimensional inference to high-dimensional cyclic factor graphs.
    \item  NF-iSAM achieves superior accuracy in comparison to state-of-the-art SLAM algorithms in describing the full posteriors encountered in highly non-Gaussian SLAM settings.
\end{enumerate}
This paper extends our conference paper \cite{huang2021nfisam} with more technical details, new SLAM problems involving data association ambiguity, and extensive experimental results including a parametric study.

\subsection*{Outline}
The rest of the paper is organized as follows. In the following section of related work (Sec.~\ref{sec: related work}), we briefly review parametric and nonparametric SLAM techniques and some recent density modeling techniques. Sec.~\ref{sec: factor graph and Bayes tree} presents the problem statement and the high-level idea of the inference framework without digging into density modeling techniques. Sec.~\ref{sec: method} delineates the formulation for modeling densities and the detailed algorithms. Sec.~\ref{sec: implementation} summarizes our implementation and experimental setups. Sec.~\ref{sec: results} provides experimental results and demonstrates the advantages of our algorithm in comparison with state-of-the-art algorithms. Finally, Sec.~\ref{sec: conclusion} concludes with a summary of the contributions of this paper and a discussion of future research directions.

\subsection*{Notation}
General notation: Deterministic values are denoted by lowercase letters while random variables are indicated by uppercase letters. If $\mathcal{V}$ is a set of indices, then $x_{\mathcal{V}}$ denotes a vector or collection of variables associated with those indices. For example, $x_{\{i,i+1,\ldots,j\}}=(x_i,x_{i+1},\ldots,x_j)$. A vector of variables with all $n$ indices is bolded, e.g., $\mathbf{x}=x_{\{1,2,\ldots,n\}}$.  Particularly, $\leftSlice{x}{d}=(x_1,x_2,\ldots,x_{d-1})$ and $\rightSlice{x}{d}=(x_{d+1},x_{d+2},\ldots,x_{n})$.  We use $p(X)$ to denote the probability density function $p_X(\boldsymbol{\cdot})$ of random variable $X$. We denote the function $p_X(x)$ at the deterministic value $x$ by $p(x)$ or $p(X=x)$. We use $m(\bigdot)$ to indicate a non-negative potential function. We use $g(\boldsymbol{\cdot};\mathbf{w})$ to indicate a function $g(\boldsymbol{\cdot})$ that is determined by parameters $\mathbf{w}$. We denote the sampling of $X$ from a distribution $p(X)$ using the notation $\sampling{x}$. The resulting $n$ i.i.d. samples are denoted by $\{x^{(k)}\}_{k=1}^{n}$. The conditional independence relation $\condIndep{X}{Y}{Z}$ reads $X$ and $Y$ are conditionally independent given $Z$.

Graphical model notation: We define a factor graph $\mathcal{G}(\mathbf{f},\latVars,\allE)$ by nodes of random variables $\latVars$ and factors $\mathbf{f}$ and edges $\allE$ between variables and factors. Variables that are adjacent to factor $f_j$ are denoted by $\latVar_{f_j}=\{\latVar_i \in \latVars | (\latVar_i,f_j)\in \allE\}$. We use $\clique$ to denote a node or clique on the Bayes tree. We also use $\clique$ to denote the collection of variables in the clique so $\clique \subseteq \latVars$. The collection of child cliques of $\clique$ is denoted by $\cClique$. The parent clique of $\clique$ is $\pClique$. The intersection of clique $\clique$ and its parent $\pClique$ is called separator $\separator=\clique \cap \pClique$ while the remaining variables in $\clique$ are called frontal variables $\frontal=\clique \setminus \pClique$.

\section{Related Work}
\label{sec: related work}
Existing methods for SLAM factor graph inference can be categorized into two classes, namely parametric and non-parametric solutions.
They possess their own strengths on tackling point and distribution estimations, respectively.


The state-of-the-art optimization-based solutions to SLAM, such as iSAM2~\cite{kaess2012isam2}, are MAP-based \emph{point} estimators that approximate the posterior distribution by a single, parametric Gaussian model. iSAM2 presents the Bayes tree~\cite{kaess2012isam2}, a graphical model that provides a probabilistic interpretation for sparse linear algebra in square root smoothing and mapping ($\sqrt{SAM}$)~\cite{dellaert2006square} and incremental smoothing and mapping (iSAM)~\cite{kaess2008isam}. The Bayes tree can be generalized to non-Gaussian settings since it is a result purely based on conditional independence structures in graphical models; however, in iSAM2, the estimated model of the joint posterior is still constructed by linear-Gaussian conditionals~\cite[Sec. 3.4]{dellaert2017factor}. Recent works of parametric SLAM solutions focus on robust MAP estimation in the presence of outliers or multi-modal factors. \cite{Rosen2021Advances} extensively reviews robust estimation techniques including switchable constraints~\cite{sunderhauf2013switchable}, robust cost functions~\cite{agarwal2013dynamic,rosen2013robust}, and mixture models~\cite{olson2013inference,pfeifer2021advancing}. However, these techniques do not aim at capturing the shape of the full posterior.


Alternatively, nonparametric models yield more expressive representations of the full posterior. These methods use sampling techniques, such as particle filters, MCMC, or nested sampling \cite{montemerlo2002fastslam,torma2010markov, huang2021reference}. The most well-known nonparametric SLAM algorithm is FastSLAM~\cite{montemerlo2002fastslam} which is based on Rao-Blackwellized particle filters.
FastSLAM leverages the relation that map features are conditionally independent once robot poses are given. However, due to sample impoverishment in particle filters, smoothed estimates degenerate as the loss of diversity in particles' paths~\cite{arulampalam2002tutorial}.
In order to further exploit conditional independence relations, a more recent method, multimodal-iSAM (mm-iSAM)~\cite{fourie2017multi}, leverages the Bayes tree~\cite{kaess2012isam2} to solve SLAM problems with a variety of non-Gaussian error sources.
mm-iSAM uses nested Gibbs sampling, derived from nonparametric belief propagation~\cite{sudderth2003nonparametric}, to approximate the \emph{marginal} belief of each node on the Bayes tree.
As a direct extension of iSAM2, MH-iSAM2~\cite{hsiao2019mh} explicitly solves point estimates of multiple modes for SLAM problems involving multiple hypotheses (e.g., uncertain data association and ambiguous loop closures), but it cannot directly tackle more general factors such as range measurements.


Posterior distribution estimation has also drawn researchers' interest in the machine learning community. Many tools such as kernel embedding \cite{gebhardt2017kernel,kim2018imitation} and Gaussian copula \cite{lafferty2012sparse,martin2020variational} have been leveraged to model non-Gaussian densities. A recent class of algorithms aims to draw samples from a non-Gaussian target distribution by estimating a transformation that maps samples from a simple reference distribution onto the target. 
These algorithms are known as transport maps \cite{el2012bayesian,parno2018transport}, or normalizing flows \cite{rezende2015variational}.
Although they have shown good performance in modeling densities, research on high-dimensional probabilistic graphical models is limited.

\section{Inference on the Bayes tree}
\label{sec: factor graph and Bayes tree}
\subsection{Factor Graphs and the Bayes tree}
\label{sec: Factor graphs for SLAM problems}
\begin{figure}[t] 
 	\centering
 	\includegraphics[width=0.99\columnwidth]{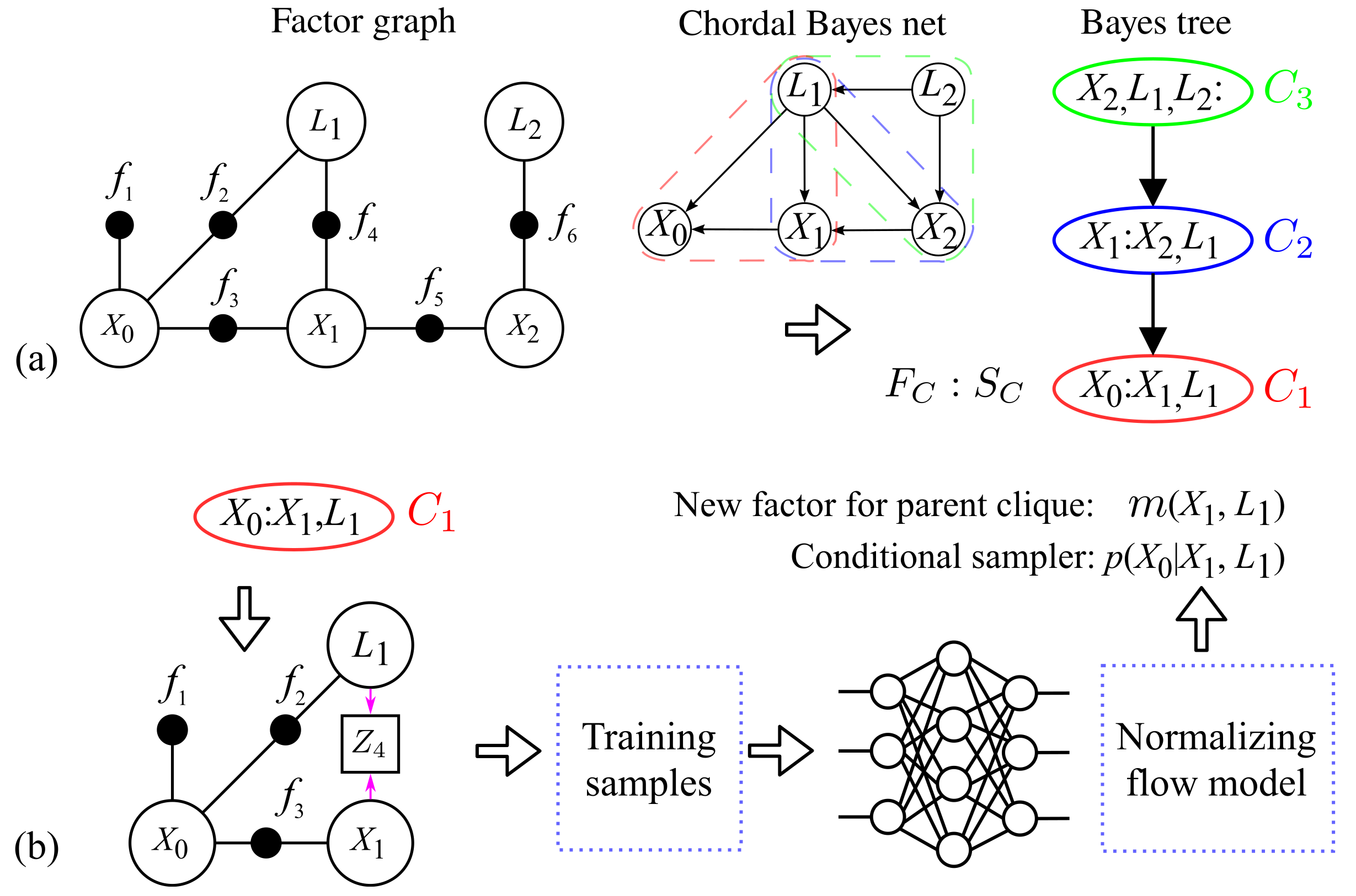}
     \caption{Illustration of core steps in NF-iSAM: (a) conversion from a factor graph to the Bayes tree with elimination ordering $(X_0, X_1, X_2, L_1, L_2)$ and (b) construction of clique conditional sampler via normalizing flows. The colon in a Bayes tree node splits frontal variables $\frontal$ and separator $\separator$. The normalizing flow model is learnt from training samples by neural networks. In factors of Bayes tree node $\clique_1$, the factor $f_4$ is reverted to a Bayes net where the measurement variable $Z_4$ is treated unobserved to enable ancestral sampling for rapidly simulating training samples (See Fig.~\ref{fig: clique normalizing flow training and learning} for more details).}
     \vspace{-0.5 cm}
     \label{fig: fig 1 illustration}
\end{figure}

We provide a brief review of factor graphs and the Bayes tree which are the foundation of our inference method. Posterior distributions in SLAM problems are usually represented by factor graphs~\cite{dellaert2017factor}, which are bipartite graphical models consisting of variable and factor nodes as shown in Fig.~\ref{fig: fig 1 illustration}a.
The state variable $\latVars:=(\latVar_1, \latVar_2,\ldots,\latVar_n)$ is a high-dimensional random variable whose $n$ components correspond to all poses and landmark locations. All measurements are denoted by $\measVals$. The posterior distribution of the factor graph is 
\begin{equation}
    p(\latVars|\measVals)=\frac{p(\measVals|\latVars)p(\latVars)}{p(\measVals)} \propto p(\measVals|\latVars)p(\latVars)= \prod_{j=1}^m f_j,\label{eqn: factor graph posterior}
\end{equation}
where $m$ is the number of factors. A factor $f_j(\latVar_{f_j})$ represents either a measurement likelihood or a prior. A prior factor has density $f_j(\latVar_{f_j})=p(\latVar_{f_j})$, where $\latVar_{f_j}$ are variables adjacent to factor $f_j$. A likelihood factor represents density $f_j(\latVar_{f_j})=p(\measVal_j|\latVar_{f_j})$ where $\measVal_j$ is the measurement in $\measVals$ that is associated with factor $f_j$. We emphasize that our objective is to infer the full posterior rather than to determine a point estimate (e.g., the MAP estimate).

\begin{figure*}[t]\vspace*{3mm}
	\centering
	\includegraphics[width=.85\textwidth]{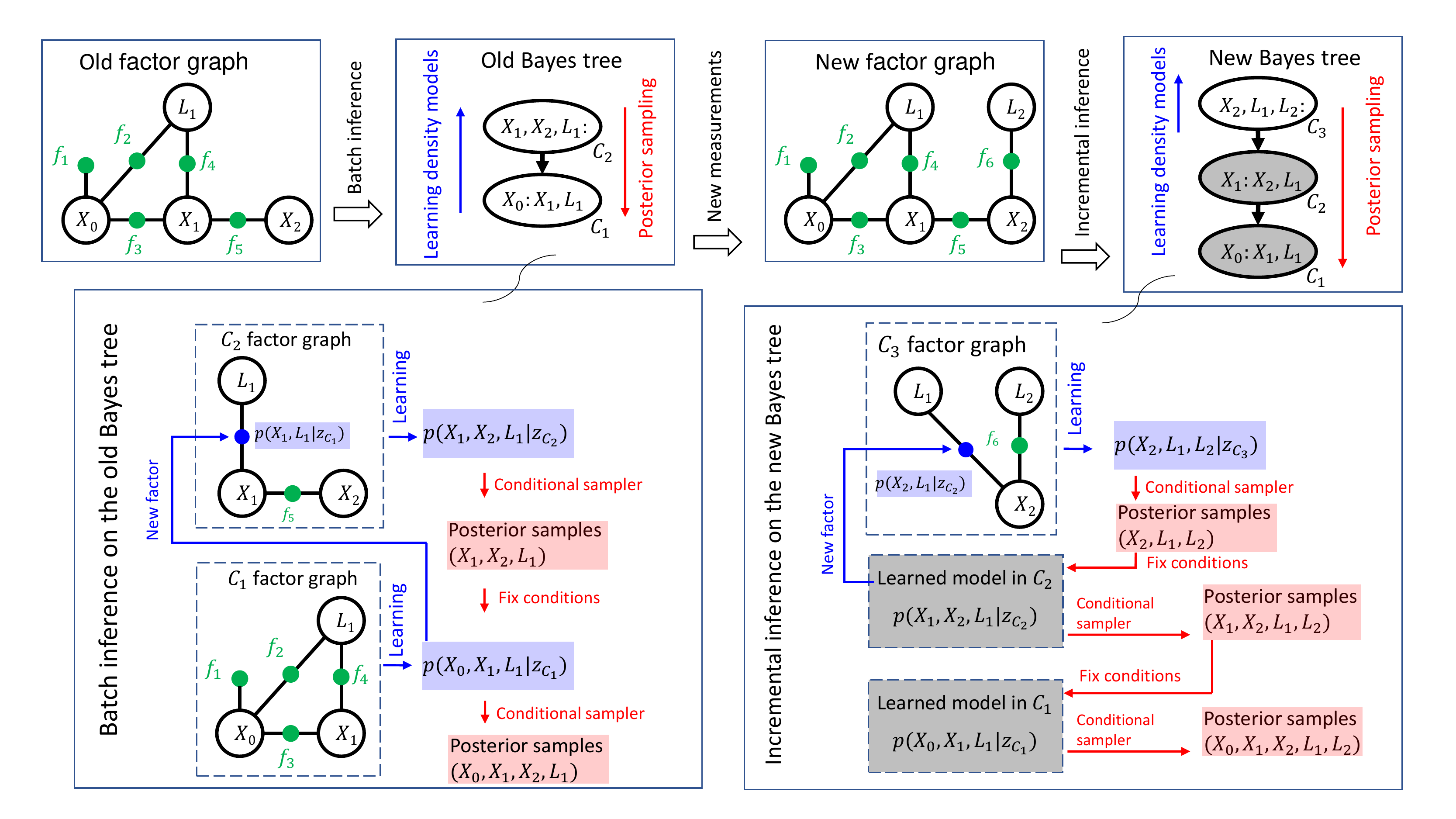}
	\vspace*{-0.05cm}
    \caption{Illustration of learning and inference procedures on Bayes trees. The variable elimination ordering $(X_0, X_1, X_2, L_1, L_2)$ for constructing Bayes trees is consistent with that in Fig.~\ref{fig: fig 1 illustration}. A Bayes tree node is denoted by $\clique$. The colon in a Bayes tree node splits frontal and separator variables. The white node on the new Bayes tree is the changes caused by new measurements. Updates of density models only occur to the changed part on the Bayes tree. Green factors in a clique factor graph are from user-defined factor graphs (e.g., old and new factor graphs here) while blue factors are separator densities from child cliques and are passed to parent cliques as new factors resulted from variable elimination.}
    \label{fig: clique factor graph}
    \vspace{-0.6cm}
\end{figure*}

The Bayes tree is a directed variant of the junction tree~\cite{dellaert2017factor}. Given a variable elimination ordering, a factor graph can be converted to a Bayes tree by the variable elimination algorithm~\cite[Alg. 2]{kaess2012isam2} and the Bayes tree algorithm~\cite[Alg. 3]{kaess2012isam2}. 
Nodes on the Bayes tree represent cliques of variables as shown in Fig.\ \ref{fig: fig 1 illustration}(a). 
Variables in a clique shared with its parent clique are called the separator while the remainder are frontal variables. 
The Bayes tree factorizes the posterior by a sequence of conditionals~\cite{dellaert2017factor, Fourie2020wafr}, as seen in
\begin{equation}
    p(\Theta|z)=\prod_{\clique\in\mathbf{C}} p(F_\clique|S_\clique,\measVals)=\prod_{\clique\in\mathbf{C}} p(F_\clique|S_\clique,z_\clique),\label{eqn: Bayes tree factorization}
\end{equation}
where $\mathbf{C}$ is the collection of cliques, $F_\clique$ denotes the set of frontal variables in clique $\clique$, $S_\clique$ denotes the separator, and $\measVal_{\clique}$ denotes the set of observations in and below clique $\clique$ on the Bayes tree (we designate the root clique as the top of the tree). The last equality in \eqref{eqn: Bayes tree factorization} is a result of applying the conditional independence relation $F_\clique \bigCI  (\measVals\setminus \measVal_{\clique})  |\separator$. Note that $\separator$ is the junction between clique $\clique$ and its parent clique. Once $S_\clique$ is fixed with a realization, the measurements above $\clique$, i.e., $\measVals\setminus \measVal_{\clique}$, will not affect $F_\clique$. Thus, $\measVals\setminus \measVal_{\clique}$ can be excluded from the condition in~(\ref{eqn: Bayes tree factorization}). The factorization \eqref{eqn: Bayes tree factorization} reflects the information-theoretical view of the Bayes tree we mentioned in Section~\ref{sec:introduction}: a decomposition of the original \emph{high-dimensional} posterior into \emph{a sequence} of \emph{low-dimensional} clique conditionals $p(F_\clique|S_\clique,z_\clique)$. We will solve a sequence of \emph{low-dimensional} density modeling problems to learn the clique conditionals. The learned conditionals in turn reconstruct the posterior, which enjoys better scalability than directly learning the \emph{high-dimensional} posterior.

\subsection{Inference Using the Clique Conditionals}
\label{sec:inference-use-conditionals}
We introduce the main idea of our inference method for modeling clique conditionals and drawing samples from the full posterior distribution. The inference method consists of two steps: (i) performing the bottom-up belief propagation on the Bayes tree for learning the clique conditionals $p(F_\clique|S_\clique, z_\clique)$ and (ii) applying ancestral sampling~\cite[Ch. 8.1.2]{bishop2006pattern} to the learned conditionals for drawing posterior samples in a top-down traversal of the Bayes tree (see the batch inference in Fig.~\ref{fig: clique factor graph} for an example of the two steps).


The primary challenge is modeling the clique density $p(F_\clique,S_\clique|z_\clique)$ and extracting the clique conditional $p(F_\clique|S_\clique, z_\clique)$ in the belief propagation. The bottom-up belief propagation is the unidirectional sum-product message passing from the leaf nodes to the root node in a tree \cite[Fig. 8.52]{bishop2006pattern} (also known as the Shafer-Shenoy algorithm~\cite{shafer1990probability} for junction trees). The clique density for any clique $\clique$, as a result of the product operation in the sum-product, is formulated by
\begin{align}
    p(\frontal,\separator|\measVal_{\clique}) &\propto \prod_{q \in \cClique} p(S_{q}|\measVal_{q}) \prod_{\substack{\latVar_{f_i}\subseteq \clique \\ \latVar_{f_i}\not\subseteq S_{q} } }  f_i(\latVar_{f_i}),\label{eqn: relation between factors and conditionals}
\end{align}
where $q$ denotes any child clique of clique $\clique$ and $\cClique$ indicates the set of the child cliques.  Relation \eqref{eqn: relation between factors and conditionals} shows that the clique density contains some user-defined factors $f(\cdot)$ and separator densities $p(S_{q}|\measVal_{q})$ which are resulted from variable elimination (i.e. the sum operation) in the child cliques. We will introduce normalizing flows in the following section to model the clique density as well as extract the clique conditional and the separator density. The separator density $p(S_{\clique}|\measVal_{\clique})$ will be passed upwards as a new factor for joining in the sum-product in the parent clique of $\clique$.

In addition, we will also use normalizing flows to construct conditional samplers for each clique conditional $p(F_\clique|S_\clique=s_\clique,z_\clique)$ such that independent samples of $F_\clique$ can be drawn once the separator is fixed by a realization $s_\clique$. These conditional samplers will be created during the belief propagation and cached for performing the ancestral sampling in the top-down traversal.

\section{SLAM via normalizing flows}
\label{sec: method}
We first briefly review normalizing flows in Section \ref{sec: normalizing flows}. We then present our novel technique for modeling and learning clique conditional $p(F_\clique|S_\clique,\measVal_{\clique})$ via normalizing flows in Section \ref{sec: clique conditional sampler}. Finally, in Section \ref{sec: incremental update} we describe our \emph{incremental} inference approach that generates joint posterior samples.
\subsection{Normalizing Flows}
\label{sec: normalizing flows}
\begin{figure}[t]\vspace*{3mm}
	\centering
	\includegraphics[width=.95\columnwidth]{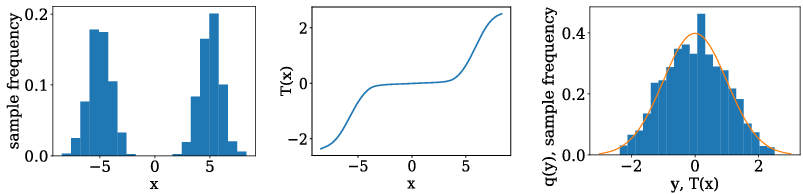}
	\vspace*{-0.05cm}
    \caption{A one-dimensional example of transformation function: histogram of sample $x$ (left), transformation function $T(x)$ (middle), and histogram of transformed samples and reference variable $y\sim N(0,1)$ (right).}
    \label{fig: 1d normalizing flow}
    \vspace{-0.6cm}
\end{figure}

Normalizing flows have shown strong expressive power for modeling complex distributions. An extensive review can be found in \cite{kobyzev2020normalizing,papamakarios2021normalizing}. A normalizing flow is a \emph{transformation} $T$ that maps a $D$-dimensional target random variable $\mathbf{X}:=(X_1,\ldots,X_D)$ onto another $D$-dimensional random variable $\mathbf{Y}:=(Y_1,\ldots,Y_D)$ that follows a reference distribution $q(\mathbf{y})$. We choose the standard multivariate
normal distribution $\mathcal{N}(0,I_D)$ as the reference distribution which is also a common choice in related literature for its advantages in computation~\cite{durkan2019neural,jaini2019sum,rezende2020normalizing,kobyzev2020normalizing,papamakarios2021normalizing}. We will explain those advantages in the later discussion of this subsection and, in particular, stress their connections with some properties of the chosen reference distribution including sampling efficiency, separability (i.e., independent components), and log-concavity. \emph{Our objective} in this subsection is to use the reference distribution, $q(\mathbf{y})$, and the transformation, $T$, to model the target distribution, $p(\mathbf{x})$; see Fig.~\ref{fig: 1d normalizing flow} for an example.

We take the transformation to be a lower-triangular map:
\begin{align} 
T(\mathbf{x})=
\begin{bmatrix*}[l]
  T_1(x_1)\\
  T_2(x_1,x_2)\\
  \vdots\\
  T_D(x_1,x_2,\hdots,x_D)\\
\end{bmatrix*}
=
\begin{bmatrix*}[c]
  y_1\\
  y_2\\
  \vdots\\
  y_D\\
\end{bmatrix*}
=
\mathbf{y},\label{eqn: change of variables} 
\end{align}
where each function $T_d$ is differentiable, bijective, and increasing with respect to $x_d$ \cite{rezende2015variational,durkan2019neural,jaini2019sum,papamakarios2021normalizing,rezende2020normalizing}. In general, a transformation between random variables of two distributions without additional constraints is not unique. It has been proven, however, that triangular maps to a standard Gaussian exist and are unique for any non-vanishing densities \cite{carlier2010knothe,bogachev2005triangular,villani2008optimal}. Theoretically, the Knothe–Rosenblatt rearrangement provides a scheme to construct the triangular map by defining $T_1$ to $T_D$ sequentially~\cite[Ch. 1]{villani2008optimal}. However, it is computationally impractical to construct the exact Knothe–Rosenblatt rearrangement for modeling a general multivariate density. Thus, in practice, many works opt to estimate such a map by seeking the optimal one among a parameterized family of triangular maps~\cite{el2012bayesian, parno2018transport, jaini2019sum}. We will follow the same practice and show how to solve for the optimal $T$. Before the optimization problem for $T$, we review \emph{three} useful properties of triangular maps that have been widely exploited for constructing the map, drawing samples, and extracting marginals and conditionals.

\emph{Property 1}: Since  $T_d$ is differentiable, lower-triangular, and increasing with respect to $x_d$, its Jacobian matrix is triangular with positive diagonals. The absolute value of Jacobian determinant is thus given by,
\begin{align}
    |{T'}(\mathbf{x})|=\prod_{d=1}^{D} \frac{\partial T_d}{\partial x_d}\label{eqn: Jacobian}.
\end{align}
For any such $T$, by change of variables, we have $p(\mathbf{x};T)=q(T(\mathbf{x})) \,\, |{T'}(\mathbf{x})|$,
where $p(\mathbf{x};T)$ denotes a density defined by $q(\mathbf{y})$ and $T$ for modeling $p(\mathbf{x})$. Thus, with the triangular structure, $p(\mathbf{x};T)$ can be expressed by
\begin{align}
    p(\mathbf{x};T)=q(T(\mathbf{x}))\prod_{d=1}^{D} \frac{\partial T_d}{\partial x_d}\label{eqn: pullback density}.
\end{align}
The density model \eqref{eqn: pullback density} can be evaluated once we can evaluate $T$ and its Jacobian. Our goal is to find $T$ that makes $p(\mathbf{x};T)$ well approximate the target density $p(\mathbf{x})$.

\emph{Property 2}: $T_{d}$ essentially models the conditional probability $p(x_d|x_{1}, \ldots,x_{d-1})$ \cite{jaini2019sum}. This idea is also referred to as autoregressive flows in literature~\cite{papamakarios2021normalizing,kobyzev2020normalizing}.
For any $d$ where $2\leq d \leq D$,
\begin{align}
    p(x_1, \ldots, x_{d-1};T) = q(y_1, \ldots, y_{d-1}) \prod_{i=1}^{d-1} \frac{\partial T_{i}}{\partial x_{i}}
\end{align}
and
\begin{align}
    p(x_1, \ldots, x_{d};T) = q(y_1, \ldots, y_{d}) \prod_{i=1}^{d} \frac{\partial T_{i}}{\partial x_{i}}.
\end{align}
Their quotient is simply
\begin{align}
    p(x_d|x_{1}, \ldots, x_{d-1};T) = q(y_d|y_1, \ldots, y_{d-1}) \frac{\partial T_{d}}{\partial x_{d}}\label{eqn: quotient}.
\end{align}
Furthermore, since we defined $q(\mathbf{y})$ as the standard multivariate normal distribution, \eqref{eqn: quotient} can be reduced to
\begin{align}
    p(x_d|x_{1}, \ldots, x_{d-1};T) = q(y_d)\frac{\partial T_{d}}{\partial x_{d}}\label{eqn: reduced quotient},
\end{align}
where $y_d = T_d(x_1,\ldots,x_d)$ and $q(y_d)$ is a one-dimensional normal distribution. Thus, $T_d$ fully determines how we model the conditional $p(x_d|x_{1}, \ldots,x_{d-1})$. This important property enables extracting marginals and conditionals once $T$ is learned. We will use this property to build desired clique conditional samplers on the Bayes tree in Section \ref{sec: clique conditional sampler}.

\emph{Property 3}: The normalizing flow $T$ provides the following simple procedure for generating samples from $p(\mathbf{x};T)$:
\begin{enumerate}
    \item   Draw samples $\mathbf{y}\sim q(\mathbf{y})$;
    \item   Solve for $\mathbf{x}$ by inverting $T$, i.e.,
    \begin{align}
    \mathbf{x}
    =
    T^{-1}(\mathbf{y})
    =
    \begin{bmatrix*}[l]
      T_{1}^{-1}(y_1)\\
      T_{2}^{-1}(y_2;x_1)\\
      \vdots\\
      T_{D}^{-1}(y_D;x_1,x_2,\hdots,x_{D-1})
    \end{bmatrix*}.\label{eqn:inverse map}
    \end{align}
\end{enumerate}
In the first step, we can directly draw samples from the standard multivariate normal distribution $q(\mathbf{y})$. Since $T$ is lower triangular, in the second step one can use a forward substitution-type approach to solve for elements of $\mathbf{x}$ one by one. We denote $x_{<d}=(x_1,x_2,\ldots,x_{d-1})$. Specifically, $x_{<d}$ is solved before $x_d$ so $x_{<d}$ can determine the one-dimensional function $T_d(\boldsymbol{\cdot}\ ;x_{<d})$ that maps $x_d$ to $y_d$. Since $T_d$ was constructed to be invertible with respect to $x_d$, $x_d$ can be solved by evaluating the inverse function of $T_d(\boldsymbol{\cdot}\ ;x_{<d})$ at $y_d$. Computation of the inverse can be efficiently done numerically or analytically, depending on the specific parameterization of $T_d$~\cite{kobyzev2020normalizing,papamakarios2021normalizing}. Thus, a sample of $\mathbf{y}$ can be transformed to a sample of $\mathbf{x}\sim p(\mathbf{x};T)$ by computing components $x_1,\ldots,x_D$ recursively.

\textbf{Optimal Normalizing Flow}: 
It remains to explain how the triangular map $T$ in \eqref{eqn: change of variables} can be obtained.
Given $n$ i.i.d.\ training samples $\{\mathbf{x}^{(k)}\}_{k=1}^n$ from  $p(\mathbf{x})$, we find $T$ by minimizing the Kullback–Leibler (KL) divergence between $p(\mathbf{x})$ and $p(\mathbf{x};T)$. In the following section (Sec. \ref{sec: clique conditional sampler}), we will present a simulation-based method for obtaining these training samples in the context of factor graph inference; in this section, we only focus on the procedure from training samples to normalizing flows in Fig.~\ref{fig: fig 1 illustration}b. Assuming training samples are given, an optimal triangular map $T^\star$ is given by,
\begin{align}
    T^{\star} &\in \argmin_{T\in \mathfrak{T}} D_\text{KL} \left( p(\mathbf{x}) \parallel p(\mathbf{x};T) \right)\\
    &= \argmin_{T\in \mathfrak{T}} \int_{\mathbf{x}}p(\mathbf{x}) \log{\frac{p(\mathbf{x})}{p(\mathbf{x};T)}}\,\text{d}\mathbf{x}\\
    &= \argmin_{T\in \mathfrak{T}} \int_{\mathbf{x}}-p(\mathbf{x}) \sum_{d=1}^{D}\left[\log{q(T_d)}+\log{\frac{\partial T_d}{\partial x_d}}\right]\,\text{d}\mathbf{x} \label{eqn: drop p of x}\\
    &\approx \argmin_{T\in \mathfrak{T}} - \sum_{k=1}^{n} \sum_{d=1}^{D}\left[\log{q(T_d)}+\log{\frac{\partial T_d}{\partial x_d}}\right]\bigg|_{\mathbf{x}=\mathbf{x}^{(k)}}\label{eqn: cost function}\\
    &= \argmin_{T\in \mathfrak{T}} \sum_{k=1}^{n} \sum_{d=1}^{D} \left( \frac{1}{2}T_{d}^{2}-\log{\frac{\partial T_d}{\partial x_d}}\right) \bigg|_{\mathbf{x}=\mathbf{x}^{(k)}},\label{eqn: MC KL}
\end{align}
where \eqref{eqn: drop p of x} follows from \eqref{eqn: reduced quotient} and also used the training samples to approximate the expectation by Monte Carlo integration. $\mathfrak{T}$ denotes all triangular maps defined by~\eqref{eqn: change of variables}. If the reference distribution is log-concave (e.g., Gaussian distributions here), the cost function from \eqref{eqn: cost function} is convex. As the feasible set is also convex, \eqref{eqn: MC KL} turns out to be a convex optimization problem~\cite[Lemma 1]{baptista2020adaptive}. Interested readers can find more discussion in~\cite{mesa2015scalable, kim2013efficient,parno2018transport}. While \eqref{eqn: MC KL} offers a theoretical pathway for finding an optimal $T$, it is not practical to solve \eqref{eqn: MC KL} directly as it requires searching among all admissible maps. To make this  practical, we limit our search to admissible maps in a parameterized family of transformations $\mathcal{F} \subset \mathfrak{T}$, 
\begin{align}
    T^{\star} &\in \argmin_{T\in \mathcal{F}} \sum_{k=1}^{n} \sum_{d=1}^{D} \left( \frac{1}{2}T_{d}^{2}(\mathbf{x}^{(k)})-\log{\frac{\partial T_d(\mathbf{x}^{(k)})}{\partial x_d}}\right) \label{eqn: parameterized problem}.
\end{align}
The parameterization choice determines whether or not the above optimization problem is convex. For example, if we let $T_1(x_1)=ax_1+b$ for a one-dimensional problem (i.e., $D=1$), we can examine derivatives of the cost function to derive a close-formed globally optimal solution under the parameterization, $a^{\star}={1}/{\sigma_{x_1}}$ and $b^{\star}=-{\bar{x}_1}/{\sigma_{x_1}}$, where $\bar{x}_1$ and $\sigma_{x_1}$ are empirical mean and standard deviation of $x_1$ samples. Obviously this affine transformation is not expressive enough for modeling non-Gaussian densities. Many flexible parameterizations of $T_d$ have been proposed including polynomial expansions~\cite{kim2013efficient, el2012bayesian}, sum-of-square polynomials~\cite{jaini2019sum}, and splines~\cite{durkan2019neural}. $T_d$ can be modeled as a one-dimensional function of $x_d$, $T_d(x_d;c_d(x_{<d}; \mathbf{w}_d))$,
where $c_d(x_{<d}; \mathbf{w}_d)$ is the so-called conditioner network~\cite{papamakarios2021normalizing}. The conditioner network with weights $\mathbf{w}_d$ takes $x_{<d}$ as the input and then outputs a set of parameters such as polynomial coefficients or spline segments that determine a differentiable, bijective, and increasing function of $x_d$ under the chosen parameterization~\cite{jaini2019sum, durkan2019neural}. However, except for special cases like Gaussian conditionals, it is difficult to analyze and model the conditioner and then solve for its weights since $T_d$ essentially encodes a high-dimensional conditional. Neural networks have been widely employed as universal functional approximators of conditioners in normalizing flows~\cite{kobyzev2020normalizing, papamakarios2021normalizing}. In practice, one optimizes over all weights $\mathbf{w}_{< D+1}$ for a solution to \eqref{eqn: parameterized problem} once a parameterization method and network configuration are designed; see Sec.~\ref{sec:neural-networks} for our parameterization method and neural network configuration. It is an active research topic to construct a parameterization method that possesses convexity for leveraging the convex problem \eqref{eqn: MC KL}. Interested readers can find further discussions in \cite{amos2017input}.

\begin{figure*}[t]
\centering
  \centering
  \includegraphics[width=.75\linewidth]{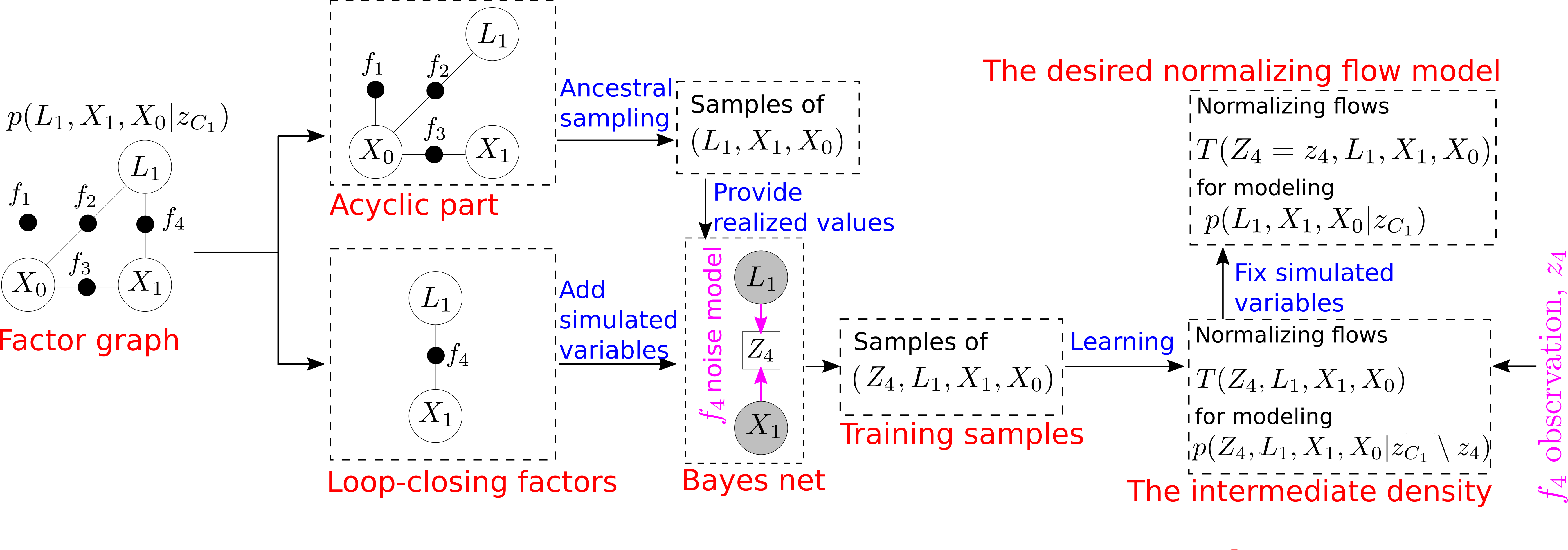}
\caption{Flowchart for the generation of training samples and the desired normalizing flow. This factor graph is the $\clique_1$ factor graph in Fig.~\ref{fig: clique factor graph}. The collection $\measVal_{\clique_1}$ denotes all measurements in this leaf clique. $\measVar_4$ and $\measVal_4$ denote the measurement variable and value in factor $f_4$, respectively.}
\label{fig: clique normalizing flow training and learning}
\end{figure*}

\emph{Practical Considerations}: A usual routine before training is standardizing raw samples by their means and standard deviations to regularize unbounded large values \cite{ioffe2015batch}. This standardizing step is equivalent to an affine transformation which makes training more efficient and does not alter the problem nor affect the non-Gaussianity in the raw samples. When the training is finished, the resulting triangular map will be transformed back to the space of raw samples by the inverse of the affine transformation. Samples of orientation variables in our SLAM experiments are transformed to $[-\pi,\pi]$ before being standardized since the experiments are treated in a planar environment; for samples in more general manifolds (e.g., $\SO(3)$), one should resort to alternative standardizing methods (e.g., in vector spaces such as $\so(3)$). Recently, more sophisticated treatments for orientation have been proposed for normalizing flows, which improves the robustness and expressive power of density estimation on complex manifolds~\cite{rezende2020normalizing}. As a widely used alternative of autoregressive flows, coupling flows impose additional structures on the triangular map at the expense of reduced expressive power, leading to improved efficiency in modeling and training normalizing flows~\cite{papamakarios2021normalizing}. While there are so many different normalizing flow parameterizations, we stress that the inference framework of NF-iSAM is generalizable as it is mostly governed by the Bayes tree and the triangular map structure. As what we will depict in the flowchart of Fig.~\ref{fig: clique normalizing flow training and learning}, learning for a model of density just takes a small fraction in the flowchart. We emphasize that the specific parameterization for modeling a complex density is a replaceable part in the pipeline. Thus, more parameterization methods can be explored and exploited in future work for efficiency, expressiveness, and robustness in density modeling.

\subsection{Clique Conditional Samplers via Normalizing Flows}
\label{sec: clique conditional sampler}
This subsection focuses on learning the clique density $p(S_\clique,F_\clique|\measVal_{\clique})$ in \eqref{eqn: relation between factors and conditionals}. We will describe in detail how to learn a normalizing flow model of the clique density from which we can extract the separator density $p(S_\clique|\measVal_{\clique})$ and the clique conditional $p(F_\clique|S_\clique,\measVal_{\clique})$.

As suggested in Section \ref{sec: normalizing flows}, we can learn the normalizing flow
\begin{equation}
    T_{\clique}(S_\clique, F_\clique)=
    \begin{bmatrix*}[l]
        T_{S_\clique}(S_\clique)\\
        T_{F_\clique}(S_\clique,F_\clique)\\
    \end{bmatrix*}
    \label{eqn: conditional normalizing flows without observations}
\end{equation}
for the clique density $p(S_\clique,F_\clique|\measVal_{\clique})$ if we have training samples from $p(S_\clique, F_\clique|\measVal_{\clique})$. Then $T_{S_\clique}$ is the normalizing flow for the separator density $p(S_\clique|\measVal_{\clique})$, and $T_{F_\clique}$ is the normalizing flow for the clique conditional $p(F_\clique|S_\clique,\measVal_{\clique})$. There are many well-developed off-the-shelf implementations of MCMC sampling such as {PyMC3} \cite{salvatier2016probabilistic} or nested sampling such as {dynesty} \cite{speagle2020dynesty}. However, even though variables in a clique are much fewer than those in the entire Bayes tree, those packages are still too slow for generating the training samples for robotics applications \cite{huang2021reference}.

Inspired by the so-called forecast-analysis scenario in hidden Markov models~\cite{spantini2019coupling} and simulating from a Bayes net model~\cite[Sec. 1.5]{dellaert2017factor}, we propose the following two-step strategy for modeling the clique density $p(F_\clique,S_\clique|\measVal_{\clique})$:
\begin{enumerate}[
    leftmargin=*,
    label={\textit{Step}\ \arabic*.},
    ref={\textit{Step}\ \arabic*}]
    \item   Draw training samples from an intermediate density $\widetilde{p}$, where sampling is efficient.\label{enu:step-1}
    \item   Train normalizing flow $\widetilde{T}$ for $\widetilde{p}$, and retrieve $T$ for $p(S_\clique,F_\clique|z_\clique)$.\label{enu:step-2}
\end{enumerate}

In \ref{enu:step-1}, we sample from the intermediate density $p(O_\clique,S_\clique,F_\clique|z_\clique')$, where $z_\clique'=z_\clique \setminus o_\clique$. We can select a set of likelihood factors, whose measurements are $o_\clique$, that breaks the factor graph of the clique into an acyclic factor graph (for example, see $f_4$ in Fig.~\ref{fig: clique normalizing flow training and learning}). We define these likelihood factors as loop-closing factors and convert them to Bayes nets where measurements are assumed as unobserved variables $O_\clique$ (see \cite[Sec. 1.7]{dellaert2017factor} for the recipe and the probabilistic interpretation of the conversion). Since both Bayes nets and the acyclic factor graph afford ancestral sampling, one can use ancestral sampling and measurement models to efficiently simulate samples of $(O_\clique,S_\clique,F_\clique)$ which are distributed according to the intermediate density.

Algorithm~\ref{algo: clique sampler} is our implementation of \ref{enu:step-1} for SLAM problems. In the algorithm, most of the loop-closing factors can be passively identified when we simulate samples. The prior factors $\mathcal{P}$ in the algorithm refer to either user-defined normalizable densities (e.g., the density of the first robot pose), from which we are typically able to draw samples directly, or separator densities modeled by normalizing flows, which enjoy fast sampling as well (\emph{Property 3}, Sec.~\ref{sec: normalizing flows}). Starting from samples in these priors (line~\ref{algo:prior-samples}), we iterate over binary factors to simulate other robot pose and landmark samples (line~\ref{algo:acyclic-factor-graph}). If both variables adjacent to a binary factor have been sampled, virtual observations between samples of these variables will be simulated (line~\ref{algo:loop-closing}). All multi-modal data association factors are proactively treated as loop-closing factors for simulating measurements (line~\ref{algo:multi-modal}).

In \ref{enu:step-2}, we use training samples from the intermediate density $p(O_\clique,S_\clique,F_\clique|\measVal_{\clique}^{\prime})$ to learn the normalizing flow $\widetilde{T}_\clique$ for eventually modeling $p(S_\clique|\measVal_{\clique})$ and $p(F_\clique|S_\clique,\measVal_{\clique})$. According to Section \ref{sec: normalizing flows}, by ordering arguments in $\widetilde{T}_\clique$ to $(O_\clique, S_\clique, F_\clique)$, we get the triangular map
\begin{equation}
    \widetilde{T}_\clique(O_\clique, S_\clique, F_\clique)=
    \begin{bmatrix*}[l]
        \widetilde{T}_{O_\clique}(O_\clique)\\
        \widetilde{T}_{S_\clique}(O_\clique,S_\clique)\\
        \widetilde{T}_{F_\clique}(O_\clique,S_\clique,F_\clique)
    \end{bmatrix*}.
    \label{eqn: conditional normalizing flows with observations}
\end{equation}
When we fix $O_\clique$ to its measured value $o_\clique$, $\widetilde{T}_{S_\clique}(O_\clique=o_\clique,S_\clique)$ gives the normalizing flow for the separator density $p(S_\clique|\measVal_{\clique})$, and $\widetilde{T}_{F_\clique}(O_\clique=o_\clique,S_\clique,F_\clique)$ gives the normalizing flow for the clique conditional $p(F_\clique|S_\clique,\measVal_{\clique})$ (see Algorithm~\ref{algo: clique normalizing flows}).
Thus, we can retrieve the desired normalizing flow model $T_\clique$ from $\widetilde{T}_\clique$:
\begin{equation}
    T_\clique\left(S_\clique,F_\clique\right)
    =
    \begin{bmatrix*}[l]
        \widetilde{T}_{S_\clique}\left(O_\clique=o_\clique,S_\clique\right) \\
        \widetilde{T}_{F_\clique}\left(O_\clique=o_\clique,S_\clique,F_\clique\right)
    \end{bmatrix*},
    \label{eqn: T tilde to T}
\end{equation}
which models the clique density $p(F_\clique,S_\clique|\measVal_{\clique})$.

\setlength\textfloatsep{.2 cm}
\begin{algorithm}[!b]
\fontsize{9pt}{9pt}\selectfont
\DontPrintSemicolon 
\KwIn{Prior $\mathcal{P}$, binary measurement $\mathcal{B}$, and multi-modal data association $\mathcal{M}$ factors in a clique}
\KwOut{Samples and measured values}
Initialize samples $\mathcal{S}$ and measured values $\mathcal{V}$ dictionaries\;

$\mathcal{S}[\latVar_i] \gets$ Sample any variable $\latVar_i$ in $\mathcal{P}$\label{algo:prior-samples}\;

\While{\text{\normalfont queue $\mathcal{B} \neq\emptyset$}}{
    $f_i=p(\measVal_i|\latVar_j, \latVar_k) \gets$ Pop the first element in $\mathcal{B}$\;
    \uIf{\text{\normalfont only one latent variable (e.g., $\latVar_k$) not in $\mathcal{S}$}}{
      $\mathcal{S}[\latVar_k] \gets$ Simulate variable $\latVar_k$ using sample $\mathcal{S}[\latVar_j]$, measured value $\measVal_i$, and measurement models \label{algo:acyclic-factor-graph}\;
    }
    \uElseIf{\text{\normalfont both latent variables $\latVar_j, \latVar_k$ in $\mathcal{S}$}}{
      $\mathcal{S}[\measVar_i] \gets$ Simulate measurement $\measVar_i$ between samples $\mathcal{S}[\latVar_j]$ and $\mathcal{S}[\latVar_k]$ using measurement models \label{algo:loop-closing}\;
      $\mathcal{V}[\measVar_i] \gets$ Measured value $\measVal_i$\;
    }
    \Else{
      Push $f_i$ to the back of $\mathcal{B}$\;
    }
}
\For{\text{\normalfont $f_i=p(\measVal_i|\latVar_{f_i})$ in $\mathcal{M}$}}{
    $\mathcal{S}[\measVar_i] \gets$ Simulate measurement $\measVar_i$ between samples $\mathcal{S}[\latVar_{f_i}]$ using measurement models\;
    $\mathcal{V}[\measVar_i] \gets$ Measured value $\measVal_i$ \label{algo:multi-modal}\;
}
\Return{\text{\normalfont Samples $\mathcal{S}$, measured values $\mathcal{V}$}}\;
\caption{{TrainingSampleSimulator}}
\label{algo: clique sampler}
\end{algorithm}

\setlength\textfloatsep{.2 cm}
\begin{algorithm}[!b]
\fontsize{9pt}{9pt}\selectfont
\DontPrintSemicolon 
\KwIn{Training samples and measured values $o$}
\KwOut{Clique conditional, separator density}
Rearrange training samples to the order of observation ($O$), separator ($S$), and frontal variables ($F$)\;
Find $\widetilde{T}$ in \eqref{eqn: conditional normalizing flows with observations} by minimizing the KL divergence according to \eqref{eqn: MC KL} using the training samples\label{algo line: minimizing KL}\;
$T(S,F) \gets \widetilde{T}(O=o,S,F)$ \tcp*{fix observations in (\ref{eqn: T tilde to T})}
$T_{S},T_{F} \gets $ partition $T(S,F)$ following (\ref{eqn: conditional normalizing flows without observations})\;
Obtain samplers of $p(F|S)$, $p(S)$ from  $T_{S}$ and $T_{F}$ by (\ref{eqn:inverse map})\;
\Return{\text{\normalfont Samplers of} $p(F|S), p(S)$}\;
\caption{{ConditionalSamplerTrainer}}
\label{algo: clique normalizing flows}
\end{algorithm}

\subsection{Incremental Inference on Bayes Tree}
\label{sec: incremental update}
Learning the full posterior distribution will start from leaf cliques ${\clique_\text{L}}$. 
By Section \ref{sec: clique conditional sampler}, we can learn normalizing flows for the separator density $p(S_{\clique_\text{L}}|\measVal_{{\clique_\text{L}}})$ and the clique conditional $p(F_{\clique_\text{L}}|S_{\clique_\text{L}},\measVal_{{\clique_\text{L}}})$. The separator density will be passed to the parent of clique ${\clique_\text{L}}$ as a new factor as shown in Fig.~\ref{fig: clique factor graph} (i.e., $p(X_1, L_1|z_{\clique_1})$ and $p(X_2, L_1|z_{\clique_2})$). The normalizing flow for the clique conditional will be saved in clique ${\clique_\text{L}}$ as a conditional sampler for sampling the joint posterior later. Our algorithm learns all normalizing flows during a single leaf-to-root traversal on the Bayes tree. The product of learned clique conditionals resolves \ref{task-1} in Sec.~\ref{sec:introduction}.

\setlength\textfloatsep{.2 cm}
\begin{algorithm}[!b]
\fontsize{9pt}{9pt}\selectfont
\DontPrintSemicolon 
\KwIn{New factors $\allFactors$, factor graph $\mathcal{G}$, ordering $\latVars$}
\KwOut{Samples of the joint posterior distribution}
$\mathcal{T} \gets \mathcal{G} \mathsf{.update(}\allFactors,\latVars\mathsf{)}$ \tcp*{update the Bayes tree}
$\mathcal{T}_\Delta \gets \mathcal{T}\mathsf{.extract(}\allFactors,\latVars\mathsf{)}$ \tcp*{extract the changed sub-tree of $\mathcal{T}$}
  \For{\text{\normalfont clique} $\clique$ \text{\normalfont in leaf-to-root traversal of} $\mathcal{T}_\Delta$}{
    $\mathbf{x}$, $o \gets$ TrainingSampleSimulator($\clique$)\;
    $p(F_\clique|S_\clique), p(S_\clique)\gets$
    ConditionalSamplerTrainer($\mathbf{x}$, $o$)\;
    Append $p(S_\clique)$ to the parent clique as a factor\;
    }
$\mathcal{D}\gets \{\}$ \tcp*{initialize an empty dictionary for posterior samples}
  \For{\text{\normalfont clique} $\clique$ \text{\normalfont in root-to-leaf traversal of} $\mathcal{T}$}{
  $p(F_\clique|S_\clique) \gets$ retrieve the conditional sampler in $\clique$ \label{algo line: retrieve conditionals}\;
  $s \gets \mathcal{D}[S_\clique]$ \label{algo line: retrieve samples} \tcp*{retrieve samples of separator}
  $\mathcal{D}[F_\clique] \gets$ draw samples from $p(\frontal|\separator=s)$ using (\ref{eqn:inverse map}) \label{algo line: draw samples}\;
}
\Return{$\mathcal{D}$}\;
\caption{{NF-iSAM}}
\label{algo: incremental infernce}
\end{algorithm}

As we described in Sec.~\ref{sec:inference-use-conditionals}, once all cliques have learned their conditional samplers of clique conditionals $p(F_{\clique}|S_{\clique},\measVal_{{\clique}})$, we can draw components of a joint posterior sample from these conditional samplers by recursively applying ancestral sampling during a root-to-leaf traversal on the Bayes tree. Through the root-to-leaf traversal, \ref{task-2} in Sec.~\ref{sec:introduction} can be accomplished. Compared to learning normalizing flows during the leaf-to-root traversal, the computational cost of the root-to-leaf traversal is minimal since normalizing flows support fast sampling (\emph{Property 3}, Sec.~\ref{sec: normalizing flows}).

When performing incremental updates, we do not need to recompute normalizing flows for all cliques of the Bayes tree. 
Every time a new factor is added into the factor graph, the corresponding change in the Bayes tree is an exact and symbolic result from the Bayes tree algorithm \cite[Alg.~3]{kaess2012isam2}. We just need to learn normalizing flows for cliques in the changed part to update posterior estimation (see clique $\clique_3$ in Fig.~\ref{fig: clique factor graph} for an example of incremental inference). 
We designate the changed part of the Bayes tree the sub-tree. The upward traversal starts from leaves of the sub-tree instead of the entire Bayes tree. 
Normalizing flows for cliques outside the sub-tree are not changed and can be reused directly. 
Thus, the computational cost for \emph{incrementally} training normalizing flows depends only on the \emph{sub-tree}, instead of the entire problem. To draw samples from the full joint posterior density, the downward traversal still needs to visit all cliques. However, as mentioned above, the computational cost for the downward sampling traversal is much lower than that for training normalizing flows. The detailed algorithm of NF-iSAM is summarized in Algorithm~\ref{algo: incremental infernce}. At this point, all \ref{task-1}-\ref{task-3} we proposed in Sec.~\ref{sec:introduction} have been resolved. With the exception of normalizing flows for modeling non-Gaussian conditionals, our strategy for incremental updates is similar to iSAM2 where linear-Gaussian conditionals  for the Gaussian approximation are partially updated as new measurements arrive. The back-substitution in iSAM2 for the least-squares solution also corresponds to a root-to-leaf traversal on the entire Bayes tree~\cite[Sec. 5.4.3]{dellaert2017factor}. 

Although the final output of our algorithms is samples for subsequent inference tasks requiring Monte Carlo integration, function evaluation of the approximate distribution can be conducted as well. While we wrap trained normalizing flows in the name of ``sampler'' in Algorithms~\ref{algo: clique normalizing flows} and~\ref{algo: incremental infernce}, the density modeled by normalizing flows can be easily evaluated by \eqref{eqn: pullback density}. Instead of drawing samples at line~\ref{algo line: retrieve samples} to line~\ref{algo line: draw samples} in Algorithm~\ref{algo: incremental infernce}, we can simply evaluate the conditionals modeled by normalizing flows and then return their product as the function evaluation of the approximate density of the joint posterior.

\section{Implementation and Experimental Setups}
\label{sec: implementation}
\subsection{Current Implementation of NF-iSAM}
\label{sec:neural-networks}
We implemented Algorithms \ref{algo: clique sampler}-\ref{algo: incremental infernce} as well as other building blocks including prior and measurement likelihood factors, factor graphs, and the Bayes tree in Python. In the current implementation of normalizing flows, we choose rational-quadratic (RQ) splines to parameterize the one-dimensional function of $x_d$, $T_d(x_d;c_d(x_{<d}; \mathbf{w}_d))$~\cite{durkan2019neural}, considering the flexibility of splines. A fully connected neural network (FCNN) with an input dimension $d-1$ is configured to model the conditioner network $c_d(x_{<d}; \mathbf{w}_d)$. If the spline is composed of $K$ different rational-quadratic functions, the conditioner network possesses an output dimension $3K-1$ of which $2K-2$ units map to coordinates of spline knots on the $x_d$-$y_d$ plane and $K+1$ units are for spline derivatives at those knots; see \emph{RQ-NSF (AR)} in \cite[Sec.\ 3]{durkan2019neural} for the detailed parameterization. Our RQ spline flows are constructed and trained using PyTorch~\cite{paszke2019pytorch}. For the experiments in Sec.~\ref{sec: results}, we use two-layer FCNNs for every conditioner network. The default number of hidden layer units in the FCNN is set to 8 and the default number of RQ splines, $K$, is set to 9. The default number of training samples is set to $2000$. While we had tried some stopping criteria in \cite{prechelt1998early} using test sets, for less training time, the training of the FCNNs stops when the loss \eqref{eqn: parameterized problem} converges. We monitor the relative change between the average loss over the latest $50$ iterations and that over the second latest $50$ iterations. The ongoing training is terminated once the relative change is lower than $1\%$.

It is worth mentioning that \cite[Table 3]{kobyzev2020normalizing} has reported that spline-based flows outperform or are on par with other normalizing flows in terms of modeling accuracy. Another advantage of RQ splines is that it can be inverted by evaluating an analytically exact expression, which permits a fast solution to the inverse transformation problem \eqref{eqn:inverse map} for sampling.

\subsection{Other Solvers and Computation Resources}
We use a nested-sampling-based approach for factor graphs, NSFG \cite{huang2021reference}, to obtain high-quality samples for some examples in Sec.~\ref{sec: results} as reference solutions. NSFG is implemented in Python based on the dynamic nested sampling package, {dynesty}\cite{speagle2020dynesty}. iSAM2 (provided by the GTSAM library in C++ \cite{dellaert2012factor}) and mm-iSAM (provided by Caesar.jl, v0.10.2 in Julia \cite{caesarjl}) are tested in our experiments as well. Experiments are performed on a workstation with an AMD Ryzen ThreadRipper 3970X CPU with 32 cores and 64 threads, an NVIDIA RTX 3090 GPUs, and 125.7 GB of RAM running Ubuntu 20.04.1 LTS. Only the neural network training in NF-iSAM uses the GPU while other solvers and other computation in NF-iSAM rely on the CPU in our experiments.

\subsection{Datasets and Measurement Likelihood Models}
\label{sec: datasets}
In the following experiments, we use three simulated datasets and two real-world datasets for range-only SLAM problems with and without data association ambiguity. We apply a unified variable elimination ordering in Algorithm~\ref{algo: incremental infernce} when solving these datasets: eliminating poses along the robot trajectory first and then landmarks. The extensive experimental results form a parameter study that investigates how the performance of NF-iSAM is affected by conditions such as: i) the magnitude of measurement noise, ii) hyperparameters in normalizing flows, iii) the fraction of factors involving data association ambiguity, iv) the randomness of robot trajectories, v) random seeds for the algorithm, and vi) the dimensionality of the SLAM problems.

We define three types of likelihood models for measurements in the datasets. First, the measurement of transformation between robot poses $T_i^{w}$ and $T_j^{w} \in \text{SE}(d)$ is modeled as $\tilde{T}_i^{j}=T_w^{j}T_i^{w}\exp{(\bm{\xi}^{\wedge})}$ where $\bm{\xi}\sim \mathcal{N}(\mathbf{0},\Sigma)$ and $T_i^{w}$ reads pose $i$ in the world frame. Second, the range measurement between a robot pose $T_i^w$ and a landmark location $\mathbf{l}_j^w$ with known data association is modeled as $\tilde{r}_i^j = \lVert \mathbf{t}_i^w - \mathbf{l}_j^w \rVert_2 + \xi $ where $\mathbf{t}_i^w$ denotes the translation vector in $T_i^w$ and $\xi \sim \mathcal{N}(0, \sigma^2)$. Third, the range measurement with unknown data association, $\tilde{r}_i$, is modeled as $p( \tilde{r}_i | \mathbf{t}_i^w, \mathcal{L}_i)=\frac{1}{|\mathcal{L}_i|} \sum_{\mathbf{l}_j^w \in \mathcal{L}_i}p( \tilde{r}_i | \mathbf{t}_i^w, \mathbf{l}_j^w)$ where $\mathcal{L}_i$ denotes the set of possibly associated landmarks. Each component in the sum-mixture is simply a likelihood model of range measurement with a given data association. All components in the mixture are equally weighted given no prior information about data associations.

\section{Results}
\label{sec: results}
\subsection{Synthetic Datasets}
\subsubsection{A Small Illustrative Example}

\setlength\dbltextfloatsep{5 pt}
\begin{figure*}[t]
\centering
  \centering
  \includegraphics[width=.85\linewidth]{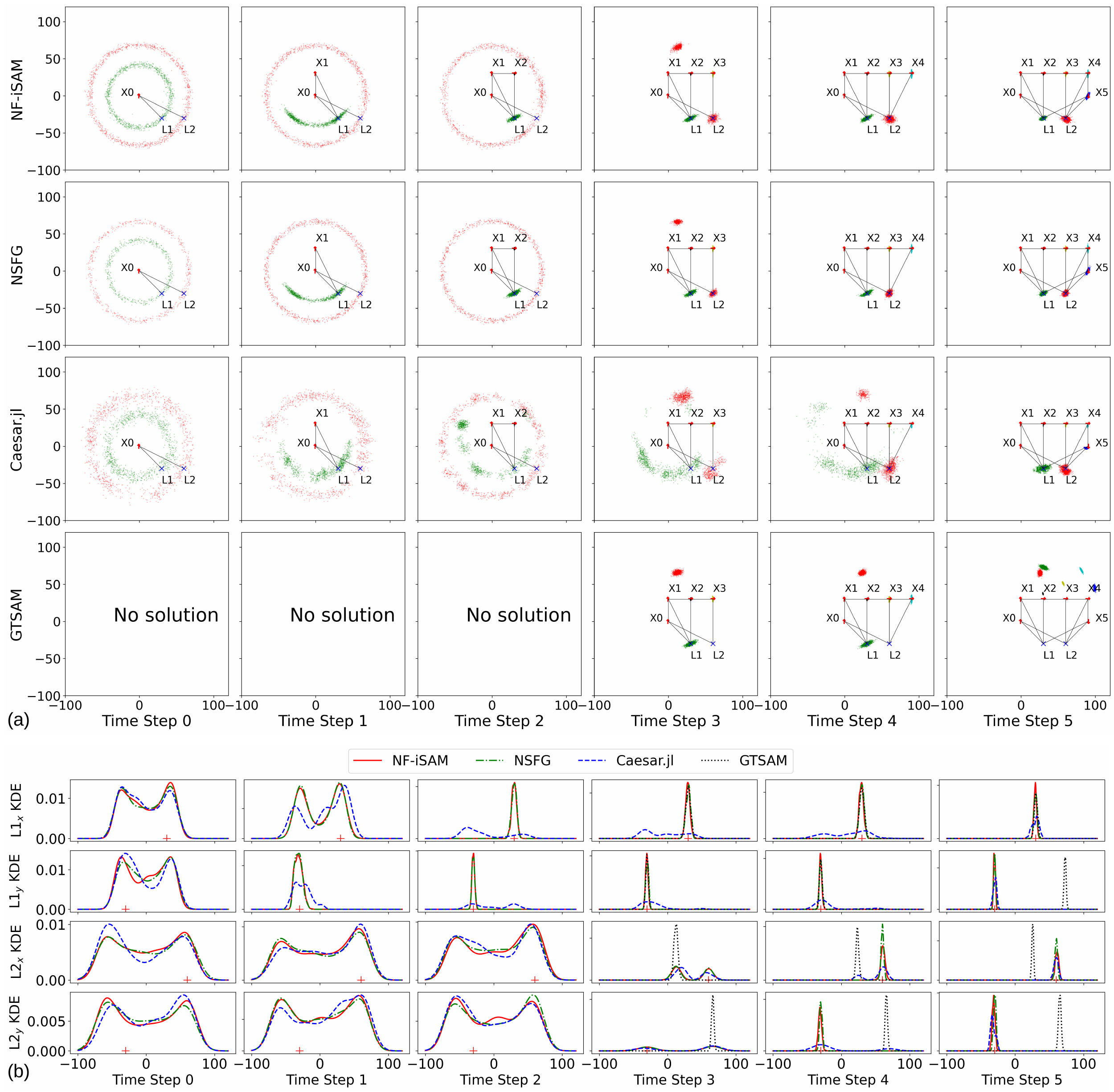}
\caption{Results for the small range-only problem without data association ambiguity: a) samples from joint posteriors and b) kernel density estimation. The robot moves clockwise from $X0$ to $X5$ and measures its distances to the landmarks $L1$ and $L2$. The black lines in (a) mark the odometry and range measurements with certain data association. Groundtruth coordinates of landmark locations are marked by `+' on KDE plots.}
\label{fig: small case with data association}
\end{figure*}

\begin{figure*}[htb]
\centering
  \centering
  \includegraphics[width=.85\linewidth]{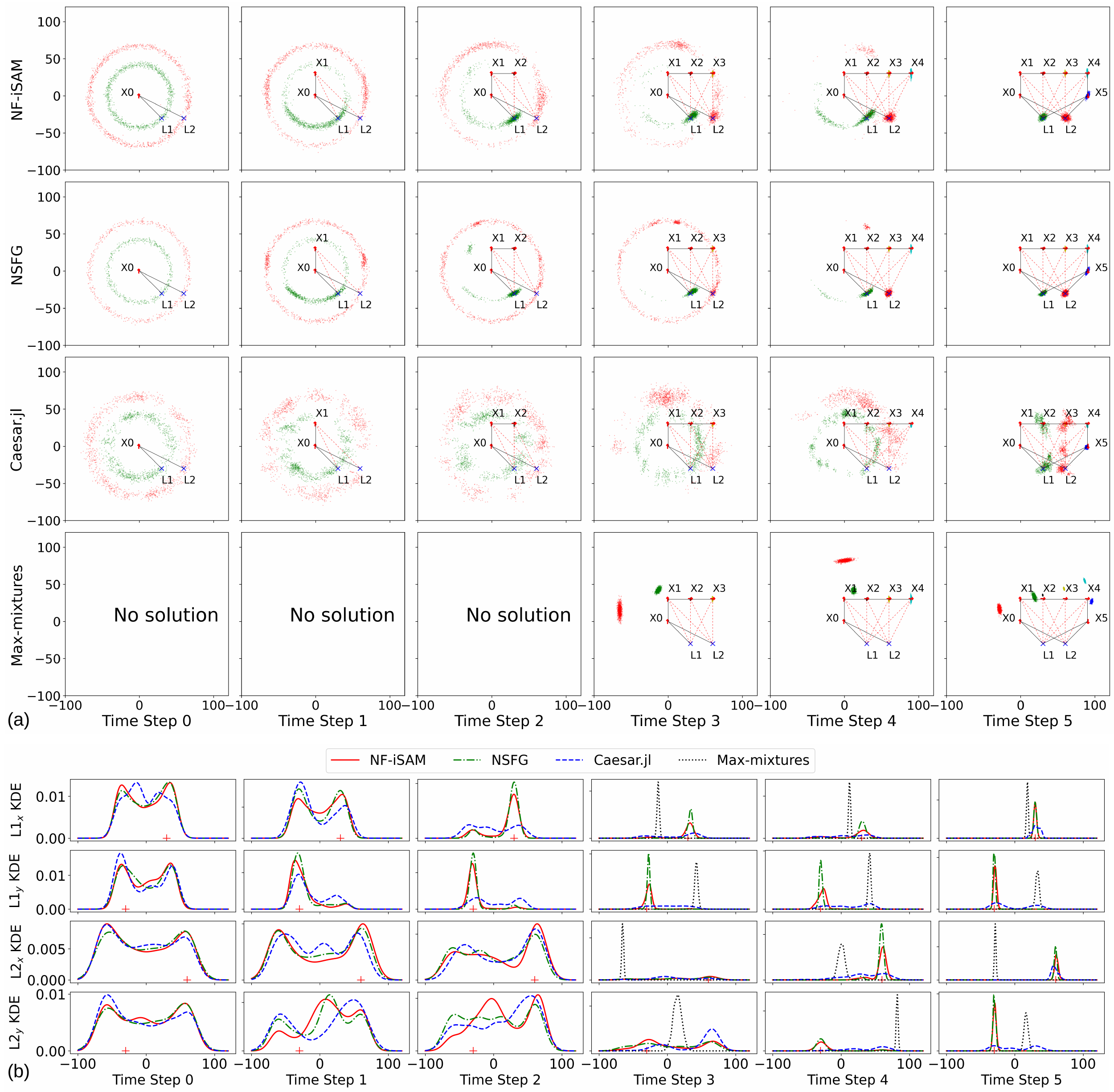}
\caption{Results for the small range-only problem with data association ambiguity: a) samples from joint posteriors and b) kernel density estimation. The robot moves clockwise from $X0$ to $X5$ and measures its distances to the landmarks $L1$ and $L2$. The black lines in (a) mark the odometry and range measurements with certain data association while the red lines in (a) between a robot pose and $k$ landmarks indicate a range measurement that is potentially associated with $k$ landmarks. Ground truth landmarks and poses in (a) are marked by `$\times$' and arrows, respectively. On the KDE plots, groundtruth landmark coordinates are marked by `+' .}
\label{fig: small case with no data association}
\end{figure*}

A small example is employed to illustrate capacities and performance of NF-iSAM on non-Gaussian inference (Fig.~\ref{fig: small case with data association} and~\ref{fig: small case with no data association}). We create a 2D environment where a robot performs a range-only SLAM task using odometry and range measurements to landmarks. A large fraction of the range measurements, however, have no identity information of landmarks, which implies that each ambiguous range measurement can (potentially) be associated with \emph{all} landmarks. While this problem is relatively low dimensional, it is nevertheless still difficult to infer the posterior distribution of robot and landmark positions since the problem involves both nonlinear measurements (e.g., distance) and high-uncertainty non-Gaussian likelihood models (due to the multi-modal data association).

\begin{figure}
\centering
  \centering
  \includegraphics[width=.8\linewidth]{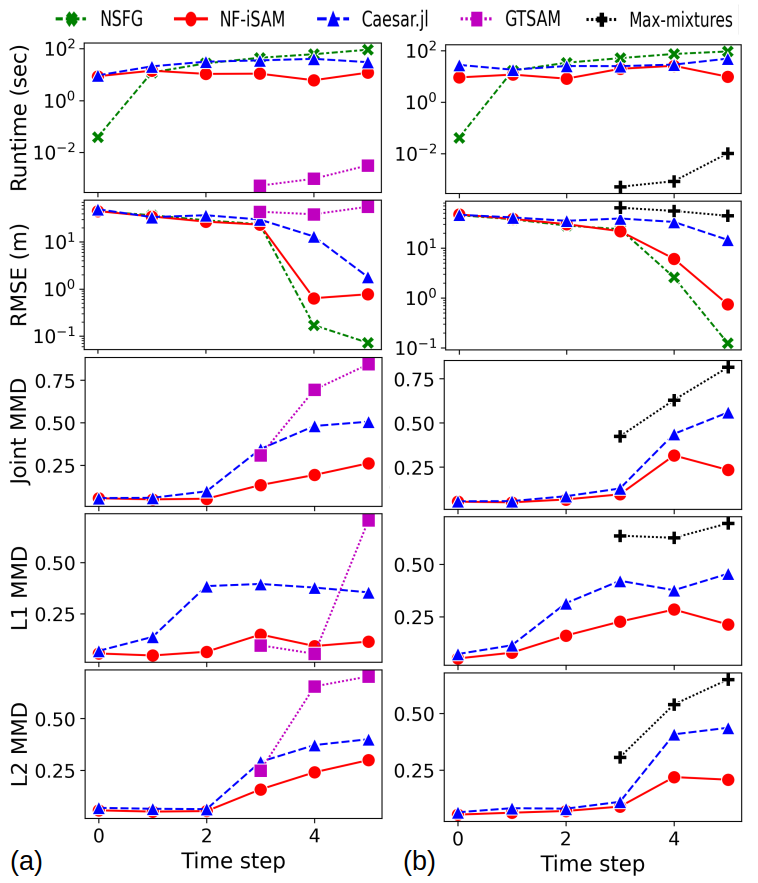}
\caption{Performance of different solvers for the small illustrative range-only problem: (a) data associations are given and (b) data associations are unknown from time step 1 to 4. The performance metrics include computation time, RMSE w.r.t. the ground truth, and maximum mean discrepancy of estimated marginal and joint posteriors to NSFG solutions.}
\label{fig: small case range pfmc}
\end{figure}

Fig.~\ref{fig: small case with data association} and~\ref{fig: small case with no data association} show samples that are drawn from estimated posteriors of problems without and with data association ambiguity. The runtime and accuracy are shown in Fig.~\ref{fig: small case range pfmc}. In both cases, the robot moves from $X_0$ to $X_5$ following a clockwise trajectory during which it acquires five odometry measurements and eight distance measurements. In the case with ambiguity, however, distance measurements from $X_{\{1-4\}}$ are modeled as potentially associated with \emph{all} detected landmarks with equal weights. Note that ``No solution" tags in the figures indicate that we could not obtain a solution due to errors thrown by GTSAM.

\setlength\dbltextfloatsep{5 pt}

Incorporating range measurements already presents a challenge due to strong nonlinearity, so we analyze the case without data association ambiguity first (see Fig.~\ref{fig: small case with data association}). According to the scatter plots and kernel density estimation, the solutions of NF-iSAM resemble reference solutions provided by NSFG for all steps. When a landmark owns two distinct distance measurements (e.g., landmark $L_1$ at time step 1 and landmark $L_2$ at time step 3), both NF-iSAM and NSFG are able to infer the bi-modal distribution of the landmark's location. Furthermore, uni-modal posterior distributions of the landmark are immediately recovered by them once the robot sights the landmark from three different poses (see landmark $L_1$ at time step 2 and landmark $L_2$ at time step 4). Caesar.jl was developed to estimate multi-modal marginal posteriors. Even though Caesar.jl pinpoints the landmarks at the last step, the uncertainty estimates are less accurate. Moreover, its estimation for earlier steps preserves many less-likely modes, which will certainly introduce errors in the evaluation of empirical mean as well as uncertainty. GTSAM leverages nonlinear least-squares (NLLS) optimization techniques to resolve Gaussian approximations of posterior distributions. For a newly detected landmark, we randomly pick a point on the circle projected by the range measurement from a robot pose, and supply it to GTSAM as the initial value of the landmark. At early steps, it cannot return a solution since the information matrix for NLLS is under-determined due to insufficient constraints. At step 3 and 4 even when each landmark possesses at least three distinct measurements, considerable drifts from the ground truth still exist as the NLLS optimization is subject to local optima in this non-convex optimization problem.

As shown in Fig.~\ref{fig: small case range pfmc}a, our quantitative analysis of this case follows the qualitative analysis above. We use the root-mean-square error (RMSE) to gauge the difference between the empirical mean of our posterior samples and the ground truth. We choose to compute the empirical mean rather than an MAP point among samples since the empirical mean can reflect errors incurred by spurious modes in estimated distributions. Here, since we aim to infer
the full posterior distribution instead of a point estimate,
\emph{maximum mean discrepancy (MMD)} \cite{gretton2012kernel} is actually a more
reasonable choice. Given samples from two densities, MMD
is a metric to evaluate how far the two distributions are apart.
Therefore, a lower MMD from a solution to the NSFG
solution indicates a more “accurate” approximation of the
posterior. As the reference solver, the RMSE of NSFG outperforms others at the expense of computation time. Note that before time step 4, the landmark belief is supposed to be bi-modal or donut-shaped distributions so it is reasonable to have large RMSE at those time steps. The plot of RMSE of NF-iSAM follows the same trend as that of NSFG and it is noticeably lower than mm-iSAM and GTSAM. We extract samples from joint and marginal distributions to compute the joint MMD and the marginal MMD respectively. The MMD plots indicate the superior  accuracy of NF-iSAM in capturing the \emph{entire} ``shape'' of the true posterior. The lower MMD of landmark $L_1$ from GTSAM at time step 3 and 4 is a coincidence as the random initial value of landmark $L_1$ happens to be around the ground truth (see green dots in scatter plots of GTSAM). However, the initial value of landmark $L_2$ unluckily locates away from the ground truth so the final estimate of GTSAM is inevitably distorted, resulting in the large MMD of landmark $L_1$ at the final time step.

\setlength\intextsep{2 pt}

\setlength\intextsep{2 pt}
\begin{figure}
\centering
  \centering
  \includegraphics[width=.8\linewidth]{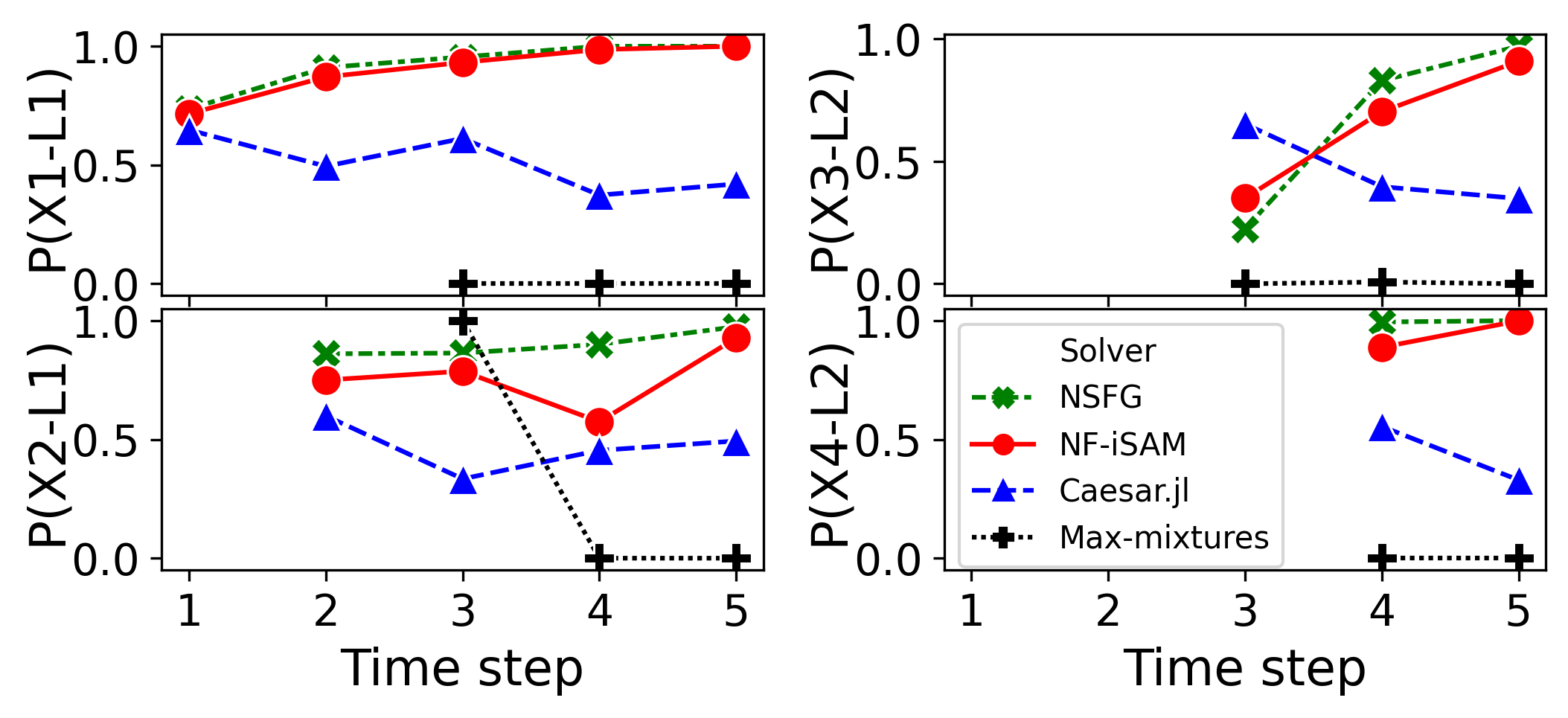}
\caption{Estimated posterior belief of groundtruth data associations for the small problem with data association ambiguity.}
\label{fig: small case data association probability}
\end{figure}

Ambiguous data association of range sensing makes the estimation problem even more difficult, as shown in the highly uncertain posteriors of NSFG solutions in Fig.~\ref{fig: small case with no data association}. At time step 2, two of the three distance measurements to landmark $L_1$ are associated with $L_2$ as well, leading to a more uncertain distribution of landmark $L_1$ than the counterpart in the ambiguity-free case. The uni-modal distributions of landmark $L_1$ and $L_2$ are not resolved until new ambiguity-free measurements added at the last step. NF-iSAM precisely captures the same trend in all scatter, kernel density estimation (KDE), and RMSE plots. The MMD plots in Fig.~\ref{fig: small case range pfmc}b indicate that NF-iSAM consistently infers more accurate estimates of the true posterior than other solvers in the setting with ambiguity. Note that iSAM2 provided by GTSAM is extended with max-mixture factors~\cite{olson2013inference, doherty2020probabilistic} for dealing with multi-modal data association so GTSAM is replaced by max-mixtures in the legend.

Given samples from the posterior distribution of the robot and landmark positions, $p(\latVars|\measVals)$, one can evaluate the posterior belief of different data associations following
\begin{align}
    p(D|\measVals) &=\int_{\latVars} \frac{p(\measVals|\latVars, D)p(D)}{\sum_{D\in \mathcal{D}} p(\measVals|\latVars, D)p(D)} p(\latVars|\measVals)\\
    &\approx \frac{1}{N}\sum_{\substack{i=1}}^{N} \frac{p(\measVals|\latVars=\latVals^{(i)}, D)p(D)}{\sum_{D\in \mathcal{D}} p(\measVals|\latVars=\latVals^{(i)}, D)p(D)}, \label{eqn: data association probability}
\end{align}
where $\latVals^{(i)}$ is one of the $N$ samples drawn from $p(\latVars|\measVals)$, and $D$ is a possible data association in the set of all associations, $\mathcal{D}$. $p(D)$ is subject to a uniform distribution over $\mathcal{D}$ as we have no prior knowledge about those associations. $p(\measVals|\latVars, D)$ is actually a binary factor under the association $D$ so it is known when we formulate the problem.

Fig.~\ref{fig: small case data association probability} shows the posterior belief of groundtruth data associations so a good estimate should approach 1 as more measurements arrive. It is clear that both NF-iSAM and NSFG manage to identify true data associations eventually.

\subsubsection{Medium-scale Problems in the Manhattan World with Range Measurements}

\begin{figure}
\centering
  \centering
  \includegraphics[width=.99\linewidth]{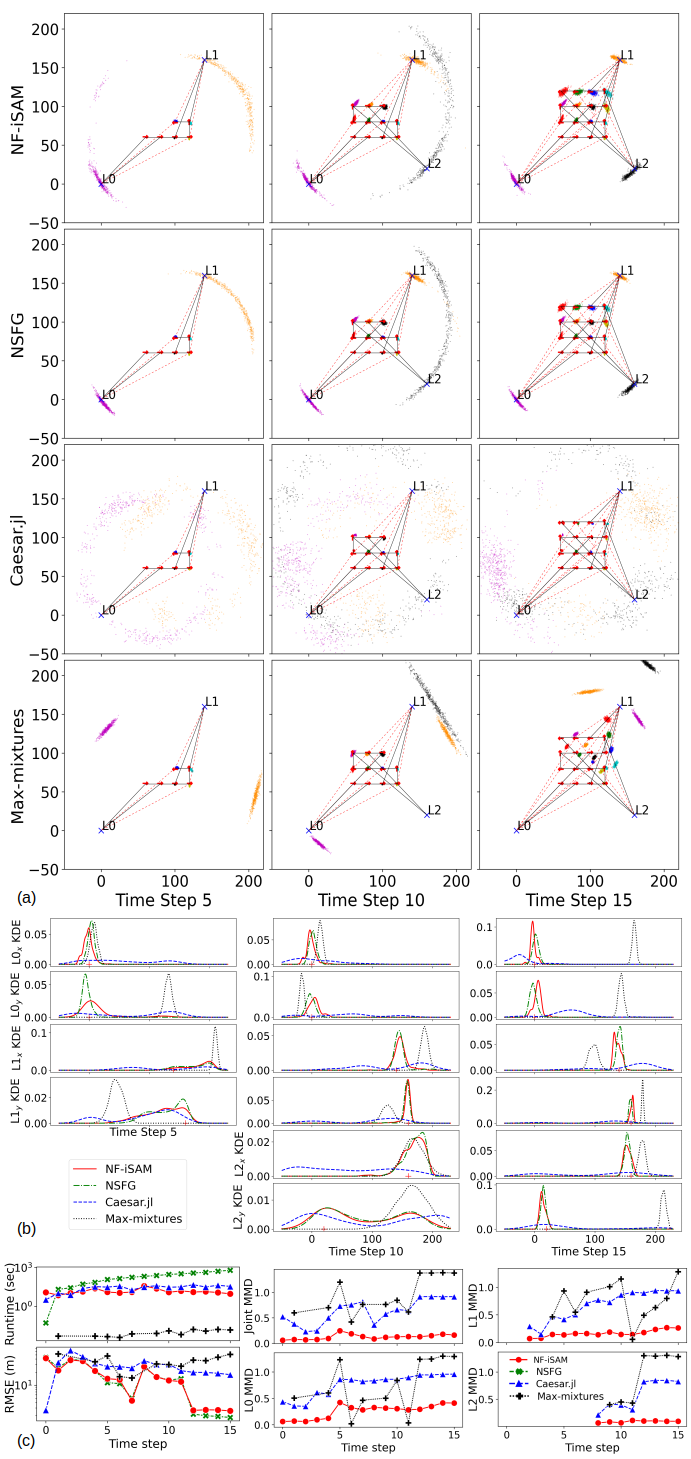}
\caption{Results for the lawnmower path problem on the 4-by-4 grid: a) samples from joint posteriors, b) kernel density estimation, and c) performance. See Fig.~\ref{fig: small case with no data association} for our convention about markers in (a).}
\label{fig: small lawnmower result sample}
\end{figure}

Here we simulate a variety of scenarios to investigate whether  NF-iSAM performs consistently well over a range of settings such as different noise magnitudes, the fraction of measurements with ambiguous data associations, and robot trajectories. We consider these to be ``medium-scale'' problems since NSFG can still converge within tens of minutes and return samples of posteriors as reference solutions for the extensive parameter study. We implement a simulator named ``Manhattan world with range measurements'' to synthesize odometry and distance measurements along Manhattan-world-like trajectories (i.e., the robot operates in grid environments where it can only move to adjacent vertices by fixed step length). Fig.~\ref{fig: small lawnmower result sample} shows a navigation task using range sensing along a lawnmower path in the simulator. Problems with random trajectories can be found in Fig.~\ref{fig: random case samples}a. In all the cases, the robot starts from time step zero and then proceeds step-by-step until time step 15, during which three landmarks will be sighted. At each time step, a distance measurement is acquired with or without data association ambiguity. Hence, unknowns at the final step consist of 16 poses and three landmark positions, resulting in a 54-dimensional posterior distribution.

\setlength\intextsep{2 pt}


There are several settings related to noise magnitudes and the fraction of ambiguous range measurements for the lawnmower path experiment. The default standard deviation of range sensing noise is set to 2 meters while the default covariance of odometry noise is $diag(0.04, 0.0016, 0.0004)$ where the diagonal entries correspond to longitudinal, lateral, and heading measurements. The default probability for generating ambiguous data association factors is set to 40\%. We are interested in investigating how those settings affect the performance of NF-iSAM and other solvers. Before varying and interrogating those settings, posterior samples for the default setting are presented in Fig.~\ref{fig: small lawnmower result sample}a to visualize the scenario and the solutions. The posterior distribution is resolved incrementally step-by-step. An interesting point in Fig.~\ref{fig: small lawnmower result sample}c is that the evidently linear curve of NSFG on the log-scale runtime plot indicates the exponential growth of computation time with increased dimensionality. On the other hand, while less accurate, other solvers are able to retain a roughly constant computation time per step by exploiting incremental inference techniques, which makes full posterior inference more tractable.

\setlength\intextsep{0 pt}
\begin{figure}
\centering
  \centering
  \includegraphics[width=.65\linewidth]{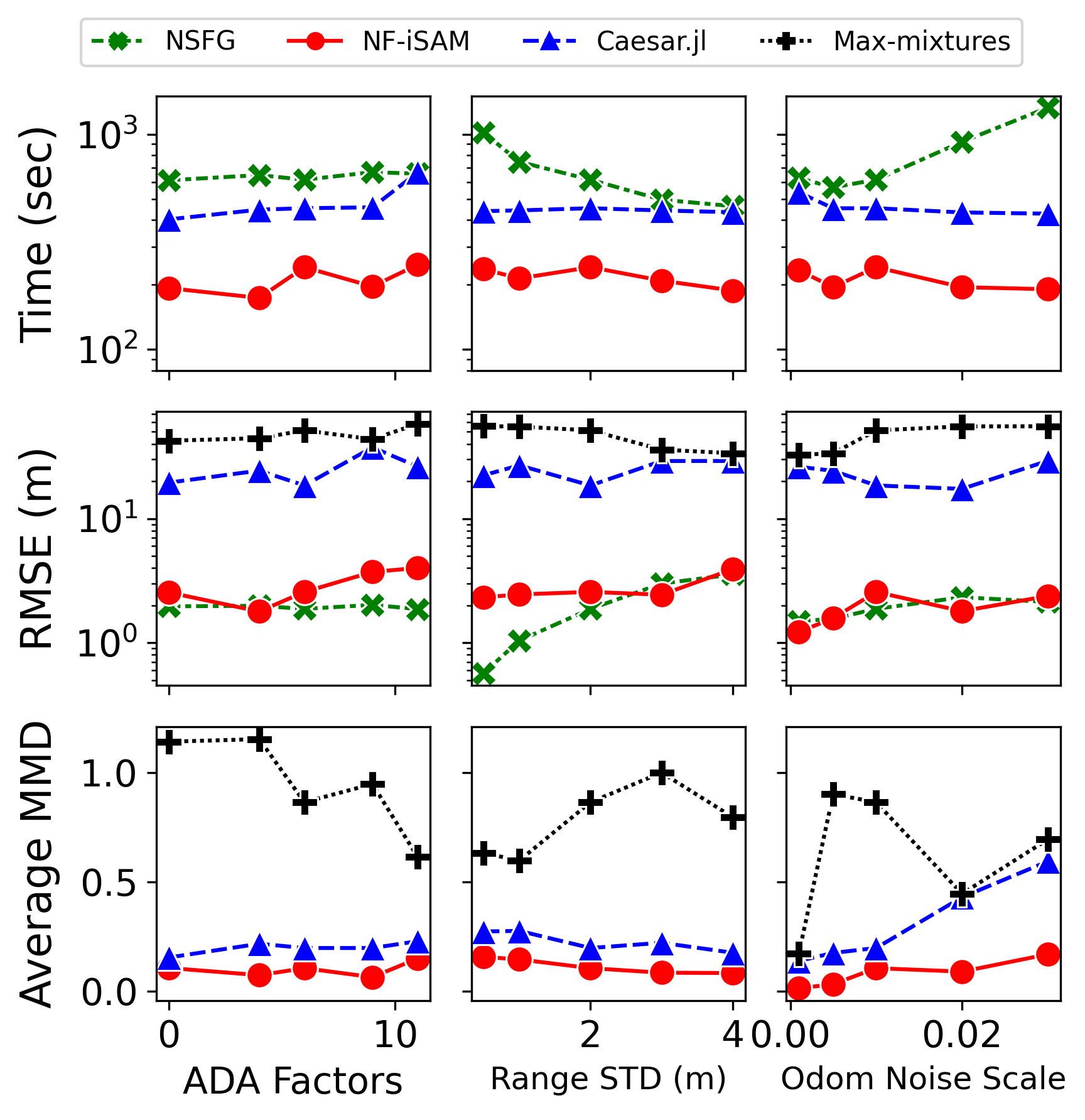}
\caption{Error and accumulated computation time for posterior estimation at the final step of the lawnmower path problem with different measurement noise and numbers of ambiguous data association factors. For each column, only one of the settings varies from the default setting. Average MMD means the average of MMDs for all estimated marginals.}
\label{fig: small lawnmower noise diversity}
\end{figure}

\setlength\intextsep{0 pt}
\begin{figure}
\centering
  \centering
  \includegraphics[width=.99\linewidth]{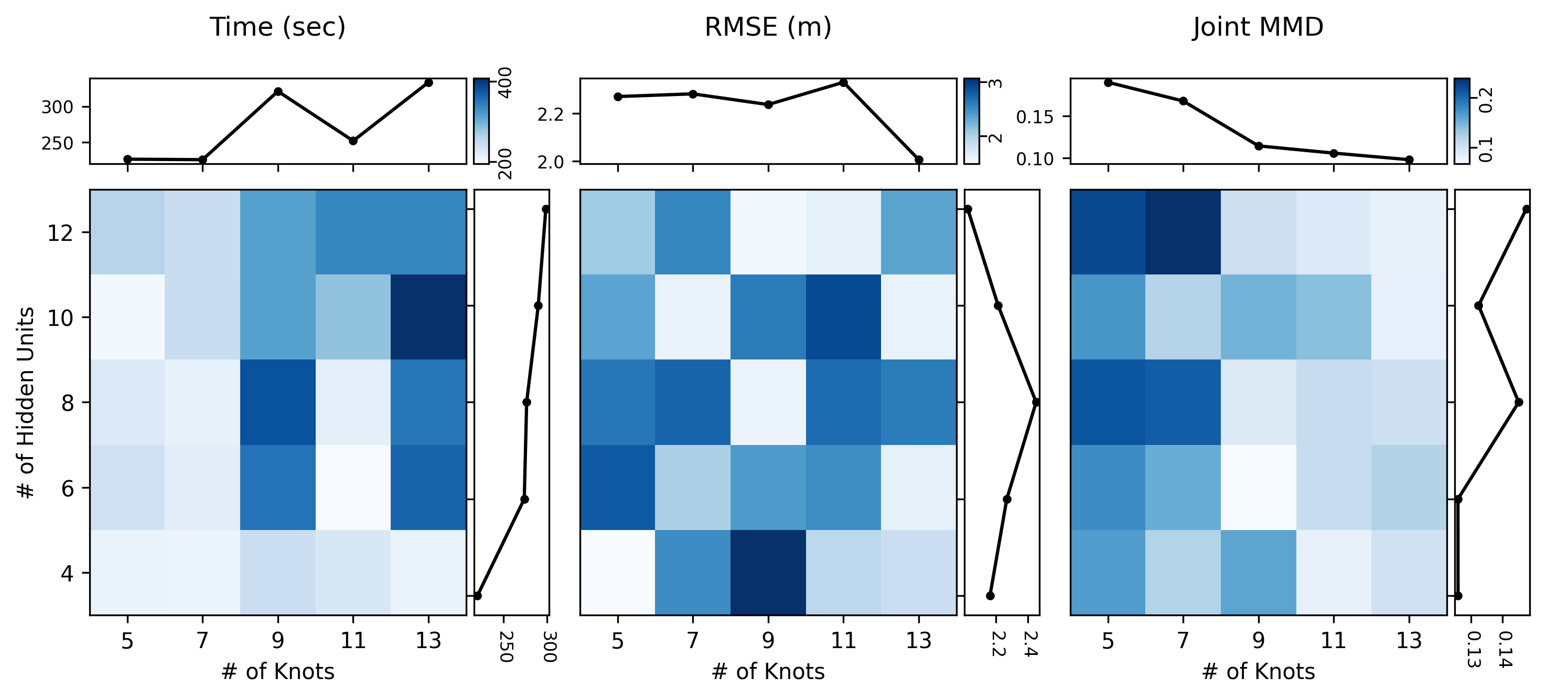}
\caption{Error and accumulated computation time of posterior estimation at the final time step with various hyper-parameters for learning normalizing flows. The parameter study is performed with the medium-scale problem with the default setting regarding noise models and data association ambiguity. Column-wise and row-wise means are shown beside the grids.}
\label{fig: small lawnmower hyperparams}
\end{figure}

We conduct an empirical study over measurement noise and the fraction of ambiguous range measurements. Only one of the settings mentioned above varies for each point in Fig.~\ref{fig: small lawnmower noise diversity}. Runtime plots of max-mixtures are neglected as they are faster than others by at least two orders of magnitude in our experiments; however, solutions of max-mixtures deviate considerably from the ground truth due to bad initial values and local optima. The computation time of NF-iSAM is consistently lower than mm-iSAM (Caesar.jl) and the reference solution (NSFG) while its RMSE is almost at the same order of magnitude as the reference solution in various settings. The lower value of average MMD from NF-iSAM indicates that its estimated posterior distribution resembles the reference posterior distribution better than mm-iSAM and GTSAM, which demonstrates the superior accuracy of NF-iSAM for full posterior estimation in non-Gaussian settings.

The normalizing flow model in NF-iSAM is primarily characterized by two predetermined hyperparameters: the number of RQ functions on the spline for fitting the one-dimensional transformation map, and the number of hidden units in the fully connected neural networks that output locations and derivatives of the spline knots (see Sec.~\ref{sec: implementation} for more about hyperparameters). The former one controls the flexibility of the spline while the latter one is the width of neural networks. Both of them have a great impact on the expressiveness of the normalizing flow model so it is worth investigating how they influence the solutions of NF-iSAM. It is not surprising to find that greater numbers of spline knots and hidden units in general lead to higher computation time as there are more parameters being trained in the neural networks (Fig.~\ref{fig: small lawnmower hyperparams}). Another considerable change is spotted in the plot of joint MMD versus the number of knots. More spline functions evidently allow for a finer fit to the shape of the posterior, decreasing the joint MMD between NF-iSAM and reference solutions. In contrast, these parameters overall present little influence on the RMSE, which implies that the enhanced expressiveness only marginally improves point estimates. The low sensitivity of RMSE also reflects our previous comment that MMD is a more reasonable choice for evaluating the quality of full posterior estimation. Furthermore, this implies an important fact that the accuracy evaluation simply using means or modes can be ineffective especially for non-Gaussian posteriors.

\begin{figure}
\centering
  \centering
  \includegraphics[width=.99\linewidth]{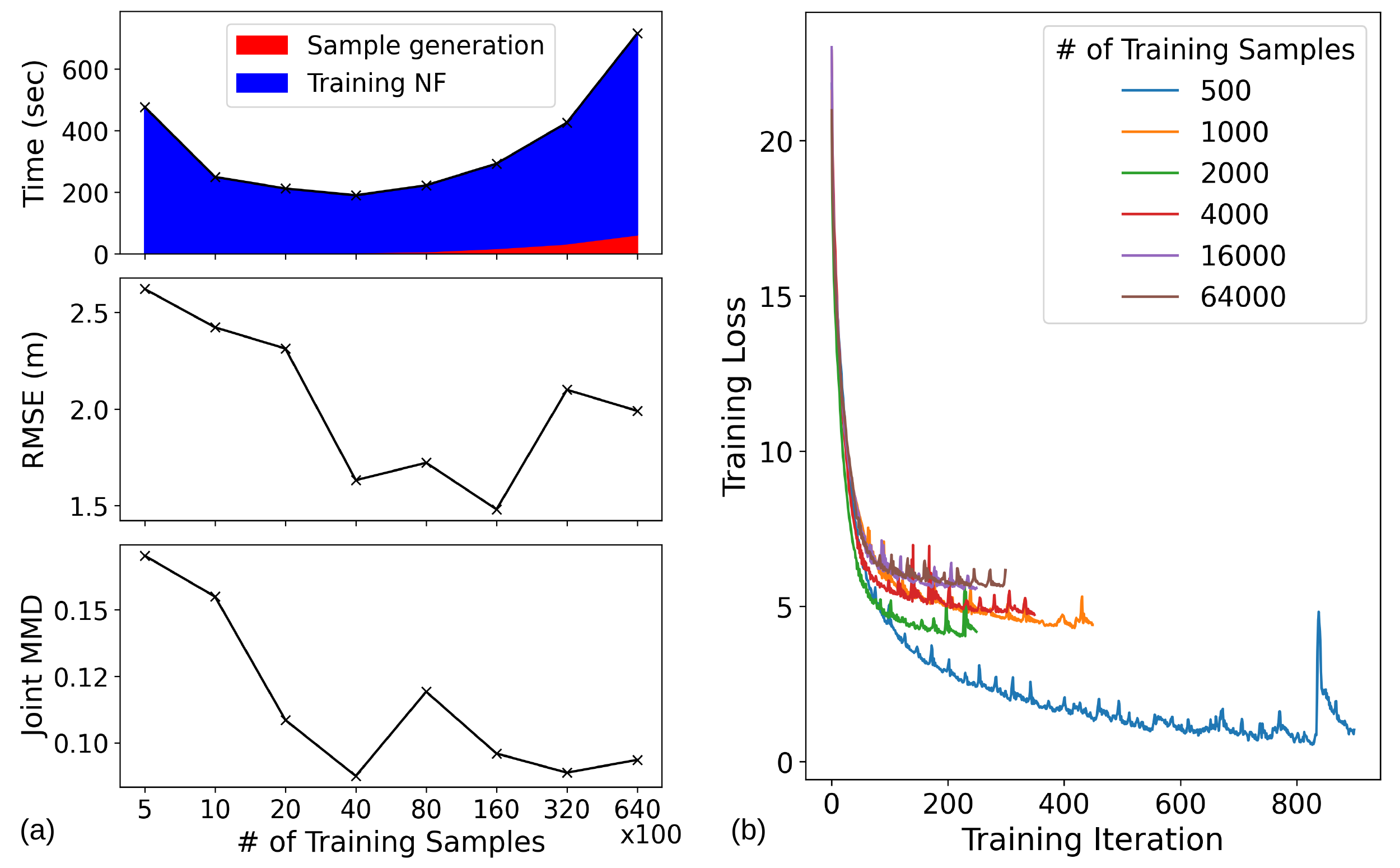}
\caption{Effects of training sample numbers in NF-iSAM results: (a) accumulated time, RMSE, and joint MMD at the final time step of the lawnmower path problem on the 4-by-4 grid, and (b) evolution of loss for training normalizing flows at the final time step.}
\label{fig: small lawnmower training samples}
\end{figure}

\begin{figure}[!h]
\centering
  \centering
  \includegraphics[width=\linewidth]{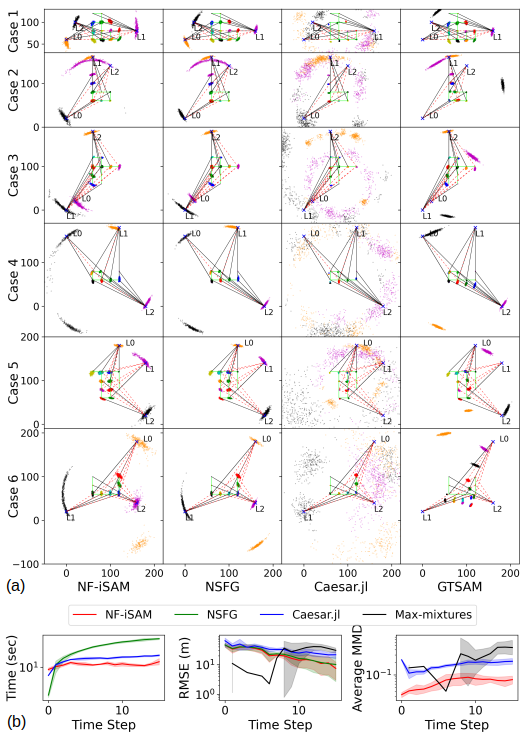}
\caption{Randomly generated cases: (a) samples of estimated posteriors and (b) error bands (95$\%$ confidence interval) of performance by different methods for randomly generated cases. Runtime of GTSAM is not shown as it makes the computation time of other solvers less distinguishable in the figure. See Fig.~\ref{fig: small case with no data association} for our convention about markers in (a).}
\label{fig: random case samples}
\end{figure}

Because training samples are self-generated in NF-iSAM, the number of training samples (i.e., $n$ in \eqref{eqn: parameterized problem}) is a predetermined hyperparameter in NF-iSAM as well. Fig.~\ref{fig: small lawnmower training samples} shows how the number of training samples affects the results for the lawnmower path problem. It is evident that small sample sizes (e.g., 500 and 1000 samples) cause less favorable performance in both computation time and accuracy. This can be explained by the plot of training loss where fewer samples incur slower and more unstable convergence of training loss. An extremely large number of training samples increases both the total computation time and the fraction of the time for sample generation; furthermore, it does not greatly improve accuracy of the results. Further investigation is needed to design an adaptive strategy that determines good hyperparameters in the training.

\setlength\intextsep{0 pt}

\setlength\intextsep{0 pt}

\setlength\intextsep{0 pt}
\begin{figure*}[!t]
\centering
  \centering
  \includegraphics[width=.99\linewidth]{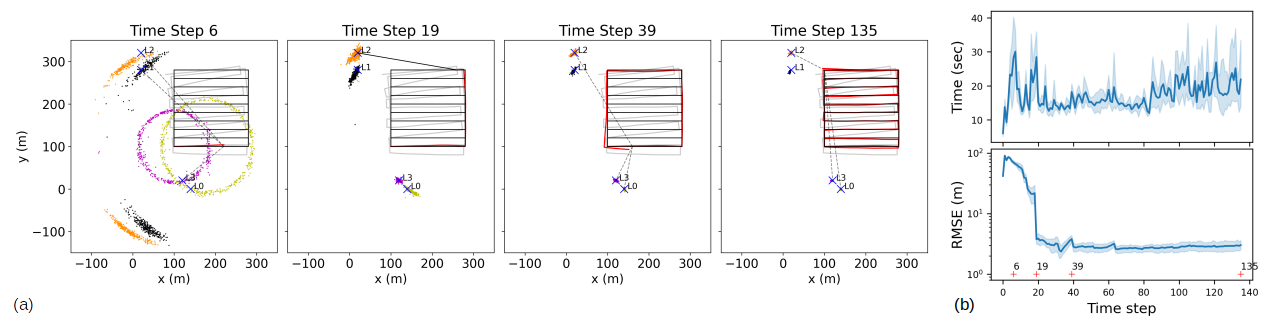}
\caption{NF-iSAM results for the simulated Plaza1 dataset: (a) posterior estimation where robot trajectories are shown in red and estimated by averages of samples, black lines and blue X markers are the ground truth, and gray lines are odometry trajectories and (b) error bands (95$\%$ confidence interval) of computation time per incremental update and RMSE of six NF-iSAM solutions to the simulated Plaza1 dataset. NF-iSAM was initialized with different random seeds to get those solutions.}
\label{fig: simulated plaza}
\end{figure*}

Fig.~\ref{fig: random case samples} presents posterior samples and performance of different algorithms under randomly generated cases. NF-iSAM behaves robustly for those cases and provides solutions almost as accurate as the reference solutions while possessing superior scalability, due to the use of incremental updates.

\subsubsection{Large-scale Problems in the Manhattan World with Range Measurements}

We demonstrate scalability of NF-iSAM and repeatability of its solutions given randomness in algorithms. We simulate a relatively large-scale range-only SLAM problem where the robot follows a path similar to that in the Plaza1 dataset~\cite{djugash2009navigating}. The entire trajectory involves 136 robot poses, 4 landmarks, 135 odometry measurements, and 136 range measurements among which 59 measurements are associated with multiple landmarks (i.e., they will be modeled by multi-modal data association factors). The posterior of robot poses and landmark locations incurs a 416-dimensional latent variable at the end of the sequence, to which a reference solution via sampling techniques is generally not available. Hence, we only compute RMSE of NF-iSAM estimates versus ground truth as the accuracy metric. 

\setlength\intextsep{0 pt}
\begin{figure*}[!t]
\centering
  \centering
  \includegraphics[width=.99\linewidth]{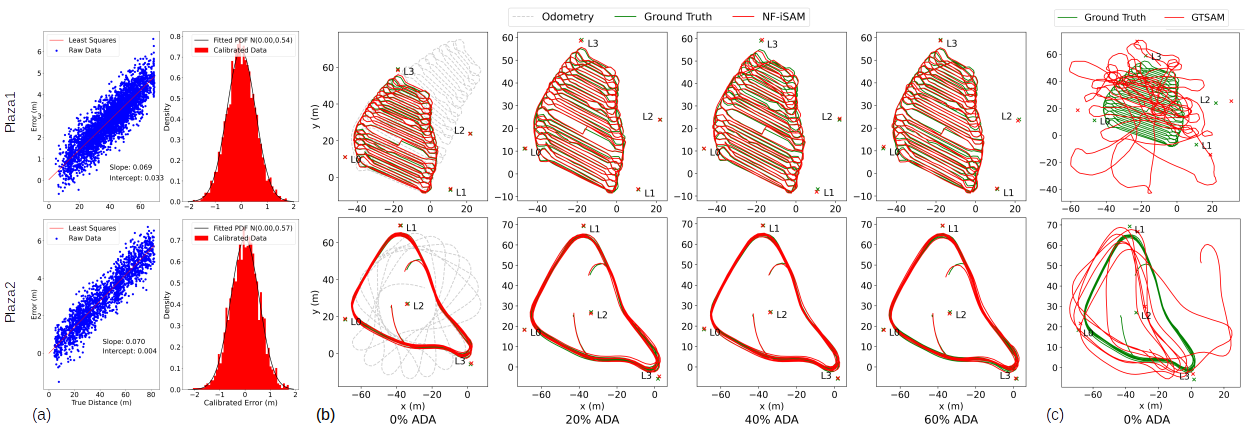}
\caption{Plaza datasets: (a) least squares for modeling distance-dependent bias in range measurements and error distributions (fitted to N(mean, standard deviation)) of the calibrated data which is obtained by subtracting least-squares-predicted bias from the raw data, (b) NF-iSAM's results for the datasets mingled with different fractions of ambiguous data association (ADA) factors, and (c) maximum a posteriori estimation by GTSAM. Trajectories by NF-iSAM are formed by the average of posterior samples. They have been processed by the Kabsch-Umeyama algorithm for trajectory alignment.}
\label{fig: real plaza traj}
\end{figure*}

\begin{figure}[!h]
\centering
  \centering
  \includegraphics[width=.99\linewidth]{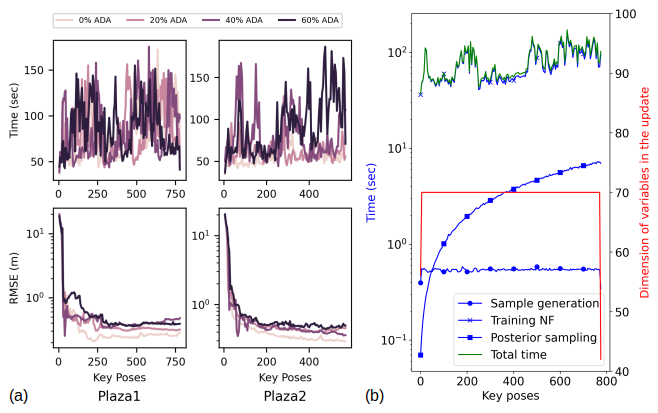}
\caption{Performance evaluation of NF-iSAM for the Plaza datasets: (a) runtime and RMSE with ADA factors and (b) decomposed runtime of NF-iSAM for the Plaza1 sequence without data association ambiguity. We also plot the dimension of variables involved in learning normalizing flows for each incremental upate.}
\label{fig: NF-iSAM plaza performance}
\end{figure}

Fig.~\ref{fig: simulated plaza}a shows the posterior samples resolved by NF-iSAM at a few important time steps. The odometry and groundtruth trajectories are respectively shown in gray and black in the figure as well. At time step 6, the robot nearly moves along a line and, as a result, the belief of landmark locations inferred by distance measurements is subject to a distribution mirrored across the line being tracked by the robot. The highly uncertain distribution of landmark location results in the significant RMSE at time step 6 as shown in Fig.~\ref{fig: simulated plaza}b. As the robot proceeds and turns left to time step 19, distance measurements acquired along the asymmetric trajectory can disambiguate landmark locations, thus the landmark distribution collapses to uni-modal. The steep decline of RMSE at time step 19 is a direct consequence of the disambiguation of landmark locations. Although the landmarks are basically pinpointed at this time step, the accumulative error in odometry can still incur inaccurate estimation. As seen around the bottom part of the groundtruth path, the odometry trajectory considerably deviates from the ground truth by more or less a block, visibly twisting the estimated trajectory. This is reflected by the sharp increase of RMSE at time step 39 as well. However, as shown in the estimated trajectory at time step 135, the deviation at the bottom of the trajectory is corrected via fusing the full-time history of measurements, which demonstrates the smoothing capacity of NF-iSAM.

In practice NF-iSAM learns probability density functions from a finite number of training samples via stochastic optimization (see Algorithms~\ref{algo: clique sampler} and \ref{algo: clique normalizing flows}), so it is not a deterministic algorithm. Therefore, it is necessary to check if NF-iSAM can achieve consistent results in the presence of inherent randomness in algorithms. As seen in Fig.~\ref{fig: simulated plaza}b, the width of the error band of RMSE is comparatively much smaller than the mean of RMSE. Thus, the accuracy of NF-iSAM is marginally affected by randomness in algorithms.

\subsection{Real-world Datasets}

We also evaluate the scalability and error of NF-iSAM using a larger real SLAM dataset.
The Plaza dataset provides time stamped range and odometry measurements ($\delta x, \delta \theta$) of a vehicle moving in a planar environment~\cite{djugash2009navigating}. Two of its sequences, {Plaza1} and {Plaza2}, are available in the {GTSAM} software distribution. There are four unknown landmarks in each of the sequences. As noted in~\cite{djugash2009navigating}, the error of distance measurement is strongly correlated with the true distance, leading to a non-zero mean in the distribution of errors. Therefore, we use least squares to fit an affine function characterizing the relation between the measurement error and the true distance. We then compute calibrated distances via subtracting the affine function from measured distances. This is a valid calibration process. If we were provided with the same sensor, we could make a collection of range measurements independently, and fit the linear model using our own data. As indicated by the histograms in Fig.~\ref{fig: real plaza traj}a, the error of the calibrated distances obeys zero-mean Gaussian distributions well and no evident outliers appear in the datasets. To tackle datasets where outliers present, the Gaussian noise model in Sec.~\ref{sec: datasets} can be replaced by outlier-robust distributions (e.g., a mixture model with a null hypothesis that the measurement is wrong~\cite{olson2013inference}). NF-iSAM is applicable to those situations since there is no assumption on noise models in our algorithms (Algorithms~\ref{algo: clique sampler}-\ref{algo: incremental infernce}).

The range-only dataset is challenging for the state-of-the-art SLAM techniques (e.g., iSAM2) that rely on the Gaussian approximation obtained by linearization around an MAP estimate. Since range-only SLAM is a highly non-convex problem and good initial values are usually not available in advance, those techniques are prone to find local optima. For a newly detected landmark, we randomly pick a point on the circle projected by the range measurement, and supply it to GTSAM as the initial value of the landmark. The GTSAM solutions in Fig.~\ref{fig: real plaza traj}c clearly show that range-only measurements pose difficulties for the MAP estimation especially when a good initialization is not available.

As shown on the leftmost side in Fig.~\ref{fig: real plaza traj}b, NF-iSAM can solve both sequences and return accurate estimates on robot trajectories and landmark positions. Moreover, we consider data association ambiguity in these datasets and evaluate the performance of NF-iSAM for multi-modal data association problems. Note that we randomly choose a faction of range measurements, wipe the ID information of the detected landmark, and designate the measurement to associate with all landmarks. The rightmost end in Fig.~\ref{fig: real plaza traj}b is the most challenging case where 60\% of range measurements are acquired with no landmark information such that they are formulated by multi-modal factors. The estimated trajectories of NF-iSAM resemble the ground truth well in all the cases with data association ambiguity. As seen in Fig.~\ref{fig: NF-iSAM plaza performance}a, although a higher fraction of data association ambiguity causes a higher RMSE, the RMSE is still at the same order of magnitude as that with no data association ambiguity. We also provide the profiling of NF-iSAM's runtime in Fig.~\ref{fig: NF-iSAM plaza performance}b. The time for sample generation and subsequent training remains roughly even across key poses since the dimension of variables on the sub-tree for incremental updates is fairly consistent. As we mentioned in Sec.~\ref{sec: incremental update}, sampling posteriors takes much shorter time than training normalizing flows. While posterior sampling is fast, it runs through the entire Bayes tree. Thus, the time for posterior sampling increases with the dimensionality of the joint posterior. To alleviate this issue for larger-scale problems, one may consider to control the size of the Bayes tree via strategies including fixed-lag smoothing\cite[Sec. 5.3.2]{dellaert2017factor}.

\section{Conclusion}
\label{sec: conclusion}
We presented a novel algorithm, NF-iSAM, that provides a promising foundation for estimating the full posterior distribution encountered in SLAM. NF-iSAM utilizes the Bayes tree coupled with normalizing flows to achieve efficient incremental updates in non-Gaussian distribution estimation of the full posterior. We demonstrated the advantages of the approach over alternative state-of-the-art point and distribution estimation techniques for SLAM, with synthetic datasets and real datasets. Our approach showed an improved estimate of the full posterior in highly non-Gaussian settings due to nonlinear measurement models and non-Gaussian (e.g., multi-modal) factors. Currently, for real SLAM problems with non-Gaussian and nonlinear models, our approach can be used to i) understand how the posterior distribution evolves over time, ii) provide reference estimates to approximate distributions found by other estimation techniques, and iii) perform inference tasks that require an estimate of the full posterior, e.g., estimating the posterior belief of data association or various expectations with respect to the posterior.

We conclude the paper by noting that NF-iSAM warrants further research as a promising and generalizable algorithmic framework. Its generalizability includes two aspects: i) the parameterization of transformation maps in normalizing flows can be replaced by other forms ii) and, more significantly, normalizing flows can be replaced by other probabilistic modeling techniques.
On the practical side, our experiments employ complex transformations to express non-Gaussian distributions, which in part incurs greater computation cost than iSAM2. Further research is needed to explore more efficient implementation strategies that can lead us closer to real-time operation, including:
(1) leveraging more efficient density modeling techniques such as coupling flows~\cite{papamakarios2021normalizing} and
(2) utilizing faster incremental update strategies on the Bayes tree such as marginalization operations and variable elimination orderings with heuristics~\cite{dellaert2017factor,Fourie2020wafr}.

\bibliographystyle{IEEEtran}
\bibliography{ref.bib}

\begin{thebibliography}{10}
\providecommand{\url}[1]{#1}
\csname url@samestyle\endcsname
\providecommand{\newblock}{\relax}
\providecommand{\bibinfo}[2]{#2}
\providecommand{\BIBentrySTDinterwordspacing}{\spaceskip=0pt\relax}
\providecommand{\BIBentryALTinterwordstretchfactor}{4}
\providecommand{\BIBentryALTinterwordspacing}{\spaceskip=\fontdimen2\font plus
\BIBentryALTinterwordstretchfactor\fontdimen3\font minus
  \fontdimen4\font\relax}
\providecommand{\BIBforeignlanguage}[2]{{%
\expandafter\ifx\csname l@#1\endcsname\relax
\typeout{** WARNING: IEEEtran.bst: No hyphenation pattern has been}%
\typeout{** loaded for the language `#1'. Using the pattern for}%
\typeout{** the default language instead.}%
\else
\language=\csname l@#1\endcsname
\fi
#2}}
\providecommand{\BIBdecl}{\relax}
\BIBdecl

\bibitem{durrant2006simultaneous}
H.~Durrant-Whyte and T.~Bailey, ``Simultaneous localization and mapping: part
  i,'' \emph{IEEE robotics \& automation magazine}, vol.~13, no.~2, pp.
  99--110, 2006.

\bibitem{Cadena2016SLAM}
C.~Cadena, L.~Carlone, H.~Carrillo, Y.~Latif, D.~Scaramuzza, J.~Neira, I.~Reid,
  and J.~Leonard, ``Past, present, and future of simultaneous localization and
  mapping: Toward the robust-perception age,'' \emph{IEEE Transactions on
  Robotics}, vol.~32, no.~6, pp. 1309--1332, Dec. 2016.

\bibitem{Rosen2021Advances}
D.~M. Rosen, K.~J. Doherty, A.~T. Espinoza, and J.~J. Leonard, ``Advances in
  inference and representation for simultaneous localization and mapping,''
  \emph{Annual Review of Control, Robotics, and Autonomous Systems}, vol.~4,
  pp. 215--242, Jan. 2021.

\bibitem{kaess2012isam2}
M.~Kaess, H.~Johannsson, R.~Roberts, V.~Ila, J.~J. Leonard, and F.~Dellaert,
  ``{iSAM2}: Incremental smoothing and mapping using the {Bayes} tree,''
  \emph{The International Journal of Robotics Research}, vol.~31, no.~2, pp.
  216--235, 2012.

\bibitem{boyd2004convex}
S.~Boyd, S.~P. Boyd, and L.~Vandenberghe, \emph{Convex optimization}.\hskip 1em
  plus 0.5em minus 0.4em\relax Cambridge university press, 2004.

\bibitem{bishop2006pattern}
C.~M. Bishop, \emph{Pattern recognition and machine learning}.\hskip 1em plus
  0.5em minus 0.4em\relax Springer, 2006, ch.~8, p. 365.

\bibitem{rosen2013robust}
D.~M. Rosen, M.~Kaess, and J.~J. Leonard, ``Robust incremental online inference
  over sparse factor graphs: Beyond the gaussian case,'' in \emph{2013 IEEE
  International Conference on Robotics and Automation}.\hskip 1em plus 0.5em
  minus 0.4em\relax IEEE, 2013, pp. 1025--1032.

\bibitem{olson2013inference}
E.~Olson and P.~Agarwal, ``Inference on networks of mixtures for robust robot
  mapping,'' \emph{The International Journal of Robotics Research}, vol.~32,
  no.~7, pp. 826--840, 2013.

\bibitem{blanco2008pure}
J.-L. Blanco, J.~Gonz{\'a}lez, and J.-A. Fern{\'a}ndez-Madrigal, ``A pure
  probabilistic approach to range-only slam,'' in \emph{2008 IEEE international
  conference on robotics and automation}.\hskip 1em plus 0.5em minus
  0.4em\relax IEEE, 2008, pp. 1436--1441.

\bibitem{long2013banana}
A.~W. Long, K.~C. Wolfe, M.~J. Mashner, and G.~S. Chirikjian, ``The banana
  distribution is {Gaussian}: A localization study with exponential
  coordinates,'' \emph{Robotics: Science and Systems VIII}, vol. 265, 2013.

\bibitem{doherty2019multimodal}
K.~Doherty, D.~Fourie, and J.~Leonard, ``Multimodal semantic slam with
  probabilistic data association,'' in \emph{2019 international conference on
  robotics and automation (ICRA)}.\hskip 1em plus 0.5em minus 0.4em\relax IEEE,
  2019, pp. 2419--2425.

\bibitem{thrun2002probabilistic}
S.~Thrun, ``Probabilistic robotics,'' \emph{Communications of the ACM},
  vol.~45, no.~3, pp. 52--57, 2002.

\bibitem{sunderhauf2013switchable}
N.~S{\"u}nderhauf, M.~Obst, S.~Lange, G.~Wanielik, and P.~Protzel, ``Switchable
  constraints and incremental smoothing for online mitigation of
  non-line-of-sight and multipath effects,'' in \emph{2013 IEEE Intelligent
  Vehicles Symposium (IV)}.\hskip 1em plus 0.5em minus 0.4em\relax IEEE, 2013,
  pp. 262--268.

\bibitem{lu2021consensus}
Z.~Lu, Q.~Huang, K.~Doherty, and J.~Leonard, ``Consensus-informed optimization
  over mixtures for ambiguity-aware object {SLAM},'' \emph{arXiv preprint
  arXiv:2107.09265}, 2021.

\bibitem{pmlr-v139-murphy21a}
\BIBentryALTinterwordspacing
K.~A. Murphy, C.~Esteves, V.~Jampani, S.~Ramalingam, and A.~Makadia,
  ``Implicit-pdf: Non-parametric representation of probability distributions on
  the rotation manifold,'' in \emph{Proceedings of the 38th International
  Conference on Machine Learning}, ser. Proceedings of Machine Learning
  Research, M.~Meila and T.~Zhang, Eds., vol. 139.\hskip 1em plus 0.5em minus
  0.4em\relax PMLR, 18--24 Jul 2021, pp. 7882--7893. [Online]. Available:
  \url{https://proceedings.mlr.press/v139/murphy21a.html}
\BIBentrySTDinterwordspacing

\bibitem{deng2021poserbpf}
X.~Deng, A.~Mousavian, Y.~Xiang, F.~Xia, T.~Bretl, and D.~Fox, ``Poserbpf: A
  rao--blackwellized particle filter for 6-d object pose tracking,'' \emph{IEEE
  Transactions on Robotics}, 2021.

\bibitem{fu2021multi}
J.~Fu, Q.~Huang, K.~Doherty, Y.~Wang, and J.~J. Leonard, ``A multi-hypothesis
  approach to pose ambiguity in object-based slam,'' in \emph{2021 IEEE/RSJ
  International Conference on Intelligent Robots and Systems (IROS)}.\hskip 1em
  plus 0.5em minus 0.4em\relax IEEE, 2021, pp. 7639--7646.

\bibitem{torma2010markov}
P.~Torma, A.~Gy{\"o}rgy, and C.~Szepesv{\'a}ri, ``A {Markov}-chain {M}onte
  {C}arlo approach to simultaneous localization and mapping,'' in
  \emph{Proceedings of the Thirteenth International Conference on Artificial
  Intelligence and Statistics}, 2010, pp. 852--859.

\bibitem{skilling2006nested}
J.~Skilling \emph{et~al.}, ``Nested sampling for general {B}ayesian
  computation,'' \emph{Bayesian analysis}, vol.~1, no.~4, pp. 833--859, 2006.

\bibitem{huang2021reference}
Q.~Huang, A.~Papalia, and J.~J. Leonard, ``Nested sampling for non-gaussian
  inference in {SLAM} factor graphs,'' \emph{arXiv preprint arXiv:2109.10871},
  2021.

\bibitem{fourie2017multi}
D.~Fourie, ``Multi-modal and inertial sensor solutions for navigation-type
  factor graphs,'' Ph.D. dissertation, Massachusetts Institute of Technology,
  2017.

\bibitem{hsiao2019mh}
M.~Hsiao and M.~Kaess, ``{MH-iSAM2}: Multi-hypothesis {iSAM} using {Bayes} tree
  and hypo-tree,'' in \emph{2019 International Conference on Robotics and
  Automation (ICRA)}.\hskip 1em plus 0.5em minus 0.4em\relax IEEE, 2019, pp.
  1274--1280.

\bibitem{Heggernes1996finding}
P.~Heggernes and P.~Matstoms, \emph{Finding good column orderings for sparse QR
  factorization}.\hskip 1em plus 0.5em minus 0.4em\relax University of
  Link{\"o}ping, Department of Mathematics, 1996.

\bibitem{Tarjan1984simple}
R.~E. Tarjan and M.~Yannakakis, ``Simple linear-time algorithms to test
  chordality of graphs, test acyclicity of hypergraphs\ , and selectively
  reduce acyclic hypergraphs,'' \emph{SIAM Journal on computing}, vol.~13,
  no.~3, pp. 566--579, 1984.

\bibitem{rezende2015variational}
D.~Rezende and S.~Mohamed, ``Variational inference with normalizing flows,'' in
  \emph{International Conference on Machine Learning}, 2015, pp. 1530--1538.

\bibitem{durkan2019neural}
C.~Durkan, A.~Bekasov, I.~Murray, and G.~Papamakarios, ``Neural spline flows,''
  in \emph{Advances in Neural Information Processing Systems}, 2019, pp.
  7511--7522.

\bibitem{jaini2019sum}
P.~Jaini, K.~A. Selby, and Y.~Yu, ``Sum-of-squares polynomial flow,'' in
  \emph{International Conference on Machine Learning}.\hskip 1em plus 0.5em
  minus 0.4em\relax PMLR, 2019, pp. 3009--3018.

\bibitem{rezende2020normalizing}
D.~J. Rezende, G.~Papamakarios, S.~Racani{\`e}re, M.~Albergo, G.~Kanwar,
  P.~Shanahan, and K.~Cranmer, ``Normalizing flows on tori and spheres,'' in
  \emph{International Conference on Machine Learning}.\hskip 1em plus 0.5em
  minus 0.4em\relax PMLR, 2020, pp. 8083--8092.

\bibitem{kobyzev2020normalizing}
I.~Kobyzev, S.~Prince, and M.~Brubaker, ``Normalizing flows: An introduction
  and review of current methods,'' \emph{IEEE Transactions on Pattern Analysis
  and Machine Intelligence}, 2020.

\bibitem{papamakarios2021normalizing}
G.~Papamakarios, E.~Nalisnick, D.~J. Rezende, S.~Mohamed, and
  B.~Lakshminarayanan, ``Normalizing flows for probabilistic modeling and
  inference,'' \emph{Journal of Machine Learning Research}, vol.~22, no.~57,
  pp. 1--64, 2021.

\bibitem{huang2021nfisam}
Q.~Huang, C.~Pu, D.~Fourie, K.~Khosoussi, J.~P. How, and J.~J. Leonard,
  ``{NF-iSAM}: Incremental smoothing and mapping via normalizing flows,''
  \emph{arXiv preprint arXiv:2105.05045}, 2021.

\bibitem{dellaert2006square}
F.~Dellaert and M.~Kaess, ``Square root sam: Simultaneous localization and
  mapping via square root information smoothing,'' \emph{The International
  Journal of Robotics Research}, vol.~25, no.~12, pp. 1181--1203, 2006.

\bibitem{kaess2008isam}
M.~Kaess, A.~Ranganathan, and F.~Dellaert, ``isam: Incremental smoothing and
  mapping,'' \emph{IEEE Transactions on Robotics}, vol.~24, no.~6, pp.
  1365--1378, 2008.

\bibitem{dellaert2017factor}
F.~Dellaert, M.~Kaess \emph{et~al.}, ``Factor graphs for robot perception,''
  \emph{Foundations and Trends in Robotics}, vol.~6, no. 1-2, pp. 1--139, 2017.

\bibitem{agarwal2013dynamic}
P.~Agarwal, G.~Tipaldi, L.~Spinello, C.~Stachniss, and W.~Burgard, ``Dynamic
  covariance scaling for robust map optimization,'' in \emph{ICRA Workshop on
  Robust and Multimodal Inference in Factor Graphs}, 2013.

\bibitem{pfeifer2021advancing}
T.~Pfeifer, S.~Lange, and P.~Protzel, ``Advancing mixture models for least
  squares optimization,'' \emph{IEEE Robotics and Automation Letters}, vol.~6,
  no.~2, pp. 3941--3948, 2021.

\bibitem{montemerlo2002fastslam}
M.~Montemerlo, S.~Thrun, D.~Koller, B.~Wegbreit \emph{et~al.}, ``{FastSLAM}: A
  factored solution to the simultaneous localization and mapping problem,''
  \emph{Aaai/iaai}, vol. 593598, 2002.

\bibitem{arulampalam2002tutorial}
M.~S. Arulampalam, S.~Maskell, N.~Gordon, and T.~Clapp, ``A tutorial on
  particle filters for online nonlinear/non-gaussian bayesian tracking,''
  \emph{IEEE Transactions on signal processing}, vol.~50, no.~2, pp. 174--188,
  2002.

\bibitem{sudderth2003nonparametric}
E.~B. Sudderth, A.~T. Ihler, W.~T. Freeman, and A.~S. Willsky, ``Nonparametric
  belief propagation,'' in \emph{Proceedings of the 2003 IEEE computer society
  conference on Computer vision and pattern recognition}, 2003, pp. 605--612.

\bibitem{gebhardt2017kernel}
G.~Gebhardt, A.~Kupcsik, and G.~Neumann, ``The kernel kalman rule—efficient
  nonparametric inference with recursive least squares,'' in \emph{Proceedings
  of the AAAI Conference on Artificial Intelligence}, vol.~31, no.~1, 2017.

\bibitem{kim2018imitation}
K.~E. Kim and H.~S. Park, ``Imitation learning via kernel mean embedding,'' in
  \emph{32nd AAAI Conference on Artificial Intelligence, AAAI 2018}.\hskip 1em
  plus 0.5em minus 0.4em\relax AAAI press, 2018, pp. 3415--3422.

\bibitem{lafferty2012sparse}
J.~Lafferty, H.~Liu, L.~Wasserman \emph{et~al.}, ``Sparse nonparametric
  graphical models,'' \emph{Statistical Science}, vol.~27, no.~4, pp. 519--537,
  2012.

\bibitem{martin2020variational}
J.~D. Martin, K.~Doherty, C.~Cyr, B.~Englot, and J.~Leonard, ``Variational
  filtering with copula models for {SLAM},'' in \emph{2020 IEEE/RSJ
  International Conference on Intelligent Robots and Systems (IROS)}.\hskip 1em
  plus 0.5em minus 0.4em\relax IEEE, 2020, pp. 5066--5073.

\bibitem{el2012bayesian}
T.~A. El~Moselhy and Y.~M. Marzouk, ``{Bayesian} inference with optimal maps,''
  \emph{Journal of Computational Physics}, vol. 231, no.~23, pp. 7815--7850,
  2012.

\bibitem{parno2018transport}
M.~D. Parno and Y.~M. Marzouk, ``Transport map accelerated {M}arkov chain
  {M}onte {C}arlo,'' \emph{SIAM/ASA Journal on Uncertainty Quantification},
  vol.~6, no.~2, pp. 645--682, 2018.

\bibitem{Fourie2020wafr}
D.~Fourie, A.~T. Espinoza, M.~Kaess, and J.~J. Leonard, ``Characterizing
  marginalization and incremental operations on the {B}ayes tree,'' in
  \emph{International Workshop on Algorithmic Foundations of Robotics (WAFR)},
  Finalnd, June 2020.

\bibitem{shafer1990probability}
G.~R. Shafer and P.~P. Shenoy, ``Probability propagation,'' \emph{Annals of
  mathematics and Artificial Intelligence}, vol.~2, no.~1, pp. 327--351, 1990.

\bibitem{carlier2010knothe}
G.~Carlier, A.~Galichon, and F.~Santambrogio, ``From {Knothe}'s transport to
  {Brenier}'s map and a continuation method for optimal transport,'' \emph{SIAM
  Journal on Mathematical Analysis}, vol.~41, no.~6, pp. 2554--2576, 2010.

\bibitem{bogachev2005triangular}
V.~I. Bogachev, A.~V. Kolesnikov, and K.~V. Medvedev, ``Triangular
  transformations of measures,'' \emph{Sbornik: Mathematics}, vol. 196, no.~3,
  p. 309, 2005.

\bibitem{villani2008optimal}
C.~Villani, \emph{Optimal transport: old and new}.\hskip 1em plus 0.5em minus
  0.4em\relax Springer Science and Business Media, 2008, vol. 338.

\bibitem{baptista2020adaptive}
R.~Baptista, O.~Zahm, and Y.~Marzouk, ``An adaptive transport framework for
  joint and conditional density estimation,'' \emph{arXiv preprint
  arXiv:2009.10303}, 2020.

\bibitem{mesa2015scalable}
D.~Mesa, S.~Kim, and T.~Coleman, ``A scalable framework to transform samples
  from one continuous distribution to another,'' in \emph{2015 IEEE
  International Symposium on Information Theory (ISIT)}.\hskip 1em plus 0.5em
  minus 0.4em\relax IEEE, 2015, pp. 676--680.

\bibitem{kim2013efficient}
S.~Kim, R.~Ma, D.~Mesa, and T.~P. Coleman, ``Efficient bayesian inference
  methods via convex optimization and optimal transport,'' in \emph{2013 IEEE
  International Symposium on Information Theory}.\hskip 1em plus 0.5em minus
  0.4em\relax IEEE, 2013, pp. 2259--2263.

\bibitem{amos2017input}
B.~Amos, L.~Xu, and J.~Z. Kolter, ``Input convex neural networks,'' in
  \emph{International Conference on Machine Learning}.\hskip 1em plus 0.5em
  minus 0.4em\relax PMLR, 2017, pp. 146--155.

\bibitem{ioffe2015batch}
S.~Ioffe and C.~Szegedy, ``Batch normalization: Accelerating deep network
  training by reducing internal covariate shift,'' in \emph{International
  conference on machine learning}.\hskip 1em plus 0.5em minus 0.4em\relax PMLR,
  2015, pp. 448--456.

\bibitem{salvatier2016probabilistic}
J.~Salvatier, T.~V. Wiecki, and C.~Fonnesbeck, ``Probabilistic programming in
  {P}ython using {PyMC3},'' \emph{PeerJ Computer Science}, vol.~2, p. e55,
  2016.

\bibitem{speagle2020dynesty}
J.~S. Speagle, ``dynesty: a dynamic nested sampling package for estimating
  {B}ayesian posteriors and evidences,'' \emph{Monthly Notices of the Royal
  Astronomical Society}, vol. 493, no.~3, pp. 3132--3158, 2020.

\bibitem{spantini2019coupling}
A.~Spantini, R.~Baptista, and Y.~Marzouk, ``Coupling techniques for nonlinear
  ensemble filtering,'' \emph{arXiv preprint arXiv:1907.00389}, 2019.

\bibitem{paszke2019pytorch}
A.~Paszke, S.~Gross, F.~Massa, A.~Lerer, J.~Bradbury, G.~Chanan, T.~Killeen,
  Z.~Lin, N.~Gimelshein, L.~Antiga \emph{et~al.}, ``Pytorch: An imperative
  style, high-performance deep learning library,'' \emph{Advances in neural
  information processing systems}, vol.~32, pp. 8026--8037, 2019.

\bibitem{prechelt1998early}
L.~Prechelt, ``Early stopping-but when?'' in \emph{Neural Networks: Tricks of
  the trade}.\hskip 1em plus 0.5em minus 0.4em\relax Springer, 1998, pp.
  55--69.

\bibitem{dellaert2012factor}
F.~Dellaert, ``Factor graphs and {GTSAM}: A hands-on introduction,'' Georgia
  Institute of Technology, Tech. Rep., 2012.

\bibitem{caesarjl}
\BIBentryALTinterwordspacing
Contributors and Dependencies, ``Caesar.jl v0.10.2,'' 2021. [Online].
  Available: \url{https://github.com/JuliaRobotics/Caesar.jl}
\BIBentrySTDinterwordspacing

\bibitem{gretton2012kernel}
A.~Gretton, K.~M. Borgwardt, M.~J. Rasch, B.~Sch{\"o}lkopf, and A.~Smola, ``A
  kernel two-sample test,'' \emph{The Journal of Machine Learning Research},
  vol.~13, no.~1, pp. 723--773, 2012.

\bibitem{doherty2020probabilistic}
K.~J. Doherty, D.~P. Baxter, E.~Schneeweiss, and J.~J. Leonard, ``Probabilistic
  data association via mixture models for robust semantic {SLAM},'' in
  \emph{2020 IEEE International Conference on Robotics and Automation
  (ICRA)}.\hskip 1em plus 0.5em minus 0.4em\relax IEEE, 2020, pp. 1098--1104.

\bibitem{djugash2009navigating}
J.~Djugash, B.~Hamner, and S.~Roth, ``Navigating with ranging radios: Five data
  sets with ground truth,'' \emph{Journal of Field Robotics}, vol.~26, no.~9,
  pp. 689--695, 2009.

\end{thebibliography}
\end{document}